\documentclass[10pt,twocolumn,letterpaper]{article}

\usepackage{cvpr}              

\usepackage{graphicx}
\usepackage{amsmath}
\usepackage{amssymb}
\usepackage{booktabs}

\usepackage{times}
\usepackage{epsfig}
\usepackage{graphicx}
\usepackage{adjustbox}
\usepackage{amsmath,amsthm,amssymb}
\newtheorem{lem}{Lemma}
\theoremstyle{remark}

\theoremstyle{remark}
\usepackage{enumerate}
\usepackage{graphicx}
\usepackage{booktabs}
\usepackage{subfiles}
\usepackage{algorithm,algpseudocode}
\makeatletter
\DeclareRobustCommand\onedot{\futurelet\@let@token\@onedot}
\def\@onedot{\ifx\@let@token.\else.\null\fi\xspace}

\def\eg{\emph{e.g}\onedot}

\def\etc{\emph{etc}\onedot}

\def\st{\emph{s.t}\onedot}

\def \x {\boldsymbol{x}}
\def \ones {\boldsymbol{1}}

\def \Real {\mathbb{R}}

\newcommand{\e}{\begin{equation}}
\newcommand{\ee}{\end{equation}}
\newcommand{\en}{\begin{equation*}}
\newcommand{\een}{\end{equation*}}
\newcommand{\eqn}{\begin{eqnarray}}
\newcommand{\eeqn}{\end{eqnarray}}
\newcommand{\bmat}{\begin{bmatrix}}
\newcommand{\emat}{\end{bmatrix}}

\newcommand{\vct}[1]{\boldsymbol{#1}}
\newcommand{\mtx}[1]{\boldsymbol{#1}}

\def \lg        {\langle}
\def \rg        {\rangle}

\newcommand{\vc}{\vct{c}}

\newcommand{\vf}{\vct{f}}

\newcommand{\vx}{\vct{x}}

\newcommand{\vdelta}{\vct{\delta}}

\newcommand{\optr}[1]{\operatorname{\textbf{tr}}(#1)}

\newcommand{\mA}{\mtx{A}}

\newcommand{\mC}{\mtx{C}}
\newcommand{\mD}{\mtx{D}}
\newcommand{\mE}{\mtx{E}}
\newcommand{\mF}{\mtx{F}}

\newcommand{\mJ}{\mtx{J}}

\newcommand{\mL}{\mtx{L}}

\newcommand{\mQ}{\mtx{Q}}

\newcommand{\mW}{\mtx{W}}
\newcommand{\mX}{\mtx{X}}
\newcommand{\mY}{\mtx{Y}}
\newcommand{\mZ}{\mtx{Z}}

\newcommand{\mSigma}{\mtx{\Sigma}}

\newcommand{\mId}{{\bf I}}

\graphicspath{{./figs/}}

\newlength{\imgwidth}
\newboolean{twoColVersion}
\setboolean{twoColVersion}{false}
\newcommand{\twoCol}[2]{\ifthenelse{\boolean{twoColVersion}} {#1} {#2} }

\usepackage[pagebackref,breaklinks,colorlinks]{hyperref}

\usepackage[capitalize]{cleveref}
\crefname{section}{Sec.}{Secs.}
\Crefname{section}{Section}{Sections}
\Crefname{table}{Table}{Tables}
\crefname{table}{Tab.}{Tabs.}

\def\cvprPaperID{5} 
\def\confName{WiCV}
\def\confYear{2022}

\begin{document}

\title{Enriched Robust Multi-View Kernel Subspace Clustering}

\author{Mengyuan Zhang\\
Clemson University\\
{\tt\small mengyuz@clemson.edu}
\and
Kai Liu\\
Clemson University\\
{\tt\small kail@clemson.edu}
}
\maketitle

\begin{abstract}
 Subspace clustering is to find underlying low-dimensional subspaces and cluster the data points correctly. In this paper, we propose a novel multi-view subspace clustering method. Most existing  methods suffer from two critical issues. First, they usually adopt a two-stage framework and isolate the processes of affinity learning, multi-view information fusion and clustering. Second, they assume the data lies in a linear subspace which may fail in practice as most real-world datasets may have non-linearity structures. To address the above issues, in this paper we propose a novel Enriched Robust Multi-View Kernel Subspace Clustering framework where the consensus affinity matrix is learned from both multi-view data and spectral clustering. Due to the objective and constraints which is difficult to optimize, we propose an iterative optimization method which is easy to implement and can yield closed solution in each step. Extensive experiments have validated the superiority of our method over state-of-the-art clustering methods.
\end{abstract}

\section{Introduction}
\label{sec:intro}
In machine learning, high-dimensional data are ubiquitous. For example, images may consist of thousands of pixels and text data may have tons of features. High dimensionality requires demanding computational time and memory, and moreover, noise in the data can bring adversely influence on performance. Fortunately, recent research shows that high-dimensional data often lies in low-dimensional structures. For instance, the set of
face images under all possible illumination conditions can be well
approximated by a 9-dimensional linear subspace \cite{basri2003lambertian}. Recovering the low-dimensional structures of data can not only save computational  cost, but also will improve the accuracy and effectiveness of learning methods. For data samples lie in low-dimensional subspaces instead of being uniformly distributed across ambient space, subspace clustering is to separate data according to their underlying subspaces and the basis for each subspace~\cite{vidal2011subspace}. For the past decade, subspace clustering has been explored
actively and applied in many applications such as image/motion/video  segmentation~\cite{yang2008unsupervised,yan2006general,goh2007segmenting}, image representation~\cite{zhu2018nonlinear,li2017structured}, \etc 

Subspace clustering approaches have been
developed and studied extensively, and among them are: iteration-based
methods such as 
\cite{tseng2000nearest,zhang2009median} which alternates cluster assignment and subspace fitting; factorization-based algebraic approaches such as \cite{kanatani2001motion,gear1998multibody,vidal2005generalized,ma2008estimation} which hypothesizes that the subspaces are independent; statistical approaches such as Multi-stage Learning \cite{gruber2004multibody}, Mixtures of Probabilistic PCA~\cite{tipping1999mixtures} which alternates between clustering and subspace estimation via Expectation Maximization; spectral clustering based approaches such as Local Subspace Affinity~\cite{yan2006general}, Locally Linear Manifold Clustering~\cite{goh2007segmenting} where data segmentation  is obtained from spectral clustering. More recently, sparse subspace clustering (SSC) has been
proposed  \cite{elhamifar2013sparse,patel2014kernel,peng2013scalable} to find a sparse representation corresponding to the data points from the same subspace. 

In the big-data era, many computer vision problems are fed with
the dataset represented by multiple feature sets, which is so called `multi-view' data.
Different descriptors characterize various and independent information from different perspectives. For instance, an
image can be described by color, texture, histogram of oriented gradients (HOG), local binary pattern (LBP),  \etc These different features can provide
useful information from different views to improve clustering performance~\cite{liu2018multiple}.
Multi-view clustering is to integrate these multiple
feature sets together to perform reliable clustering. Most existing multi-view subspace clustering methods integrate multi-view information in similarity or representation by merging multiple graphs or representation matrices into a shared one. For example, \cite{guo2013convex,tang2018learning} learn
a shared sparse subspace representation by performing matrix factorization. Similarly, centroid-based multi-view
low-rank sparse subspace clustering methods \cite{brbic2018multi,luo2018consistent,zhang2017latent} induce
low-rank and sparsity constraints on the shared affinity
matrix across different views. Instead of obtaining
a shared representation directly, Hilbert-Schmidt Independence Criterion (HSIC) and
Markov chain are introduced to learn complementary subspace representations, followed by adding them together appropriately \cite{xia2014robust,cao2015diversity}.
\begin{figure}[h!]
	\centering
	\includegraphics[width=\linewidth]{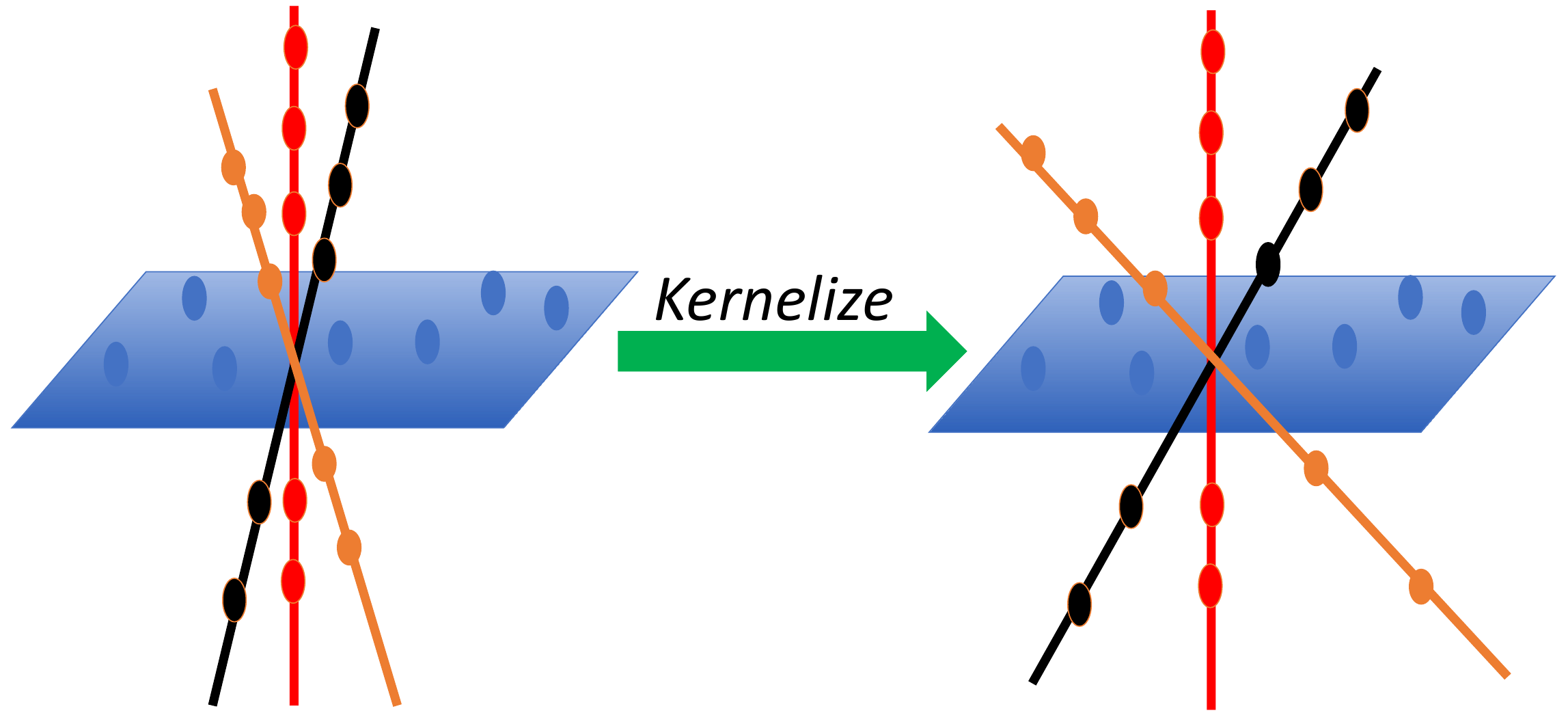}
	\caption{Using the kernel trick, data is transformed onto a high dimensional feature space so that better sparse representations can be found for subspace clustering.}
	\label{fig:kernel}
\end{figure}

Although these subspace
methods mentioned above have achieved significant success, they still suffer from the following
issues: 1) They assume the data is well separable in linear subspace, which may not be true and extensive researches show data can be better separated by mapping to higher dimension~\cite{gao2015multi,kang2020large}. 2) Previous approaches obtain the consensus affinity matrix by minimizing the squared Frobenius norm of its difference to each view, which may yield poor representation when a certain view is not well learned~\cite{zhang2020consensus,zhang2020one}. 3) Most existing approaches are usually conducted in a two-step fashion~\cite{li2015structured,zheng2020feature,heckel2013subspace}, which may fail to obtain optimal clustering
performance since the learning stage is separated from the subsequent clustering stage. The main contributions of this work are summarized as follows:
\begin{itemize}
	\item To explore the nonlinear relationship in the data, we transform the data from original space to a
	kernel space, which improves the performance of multi-view
	subspace clustering when dealing with non-linearity (\eg. specific manifold) data distributions, as shown in Figure \ref{fig:kernel}.
	\item We investigate a unified multi-view subspace clustering
	framework which jointly optimizes similarity learning and spectral clustering, where enriched consensus affinity matrix is learned from different sources.
	\item We provide a formulation to obtain robust consensus affinity matrix, which yields better clustering result.
	\item We propose an updating algorithm with closed solution in each step, which is computationally efficient. 
\end{itemize}

\begin{figure*}[t!]
	\centering
		\includegraphics[width=.8\linewidth]{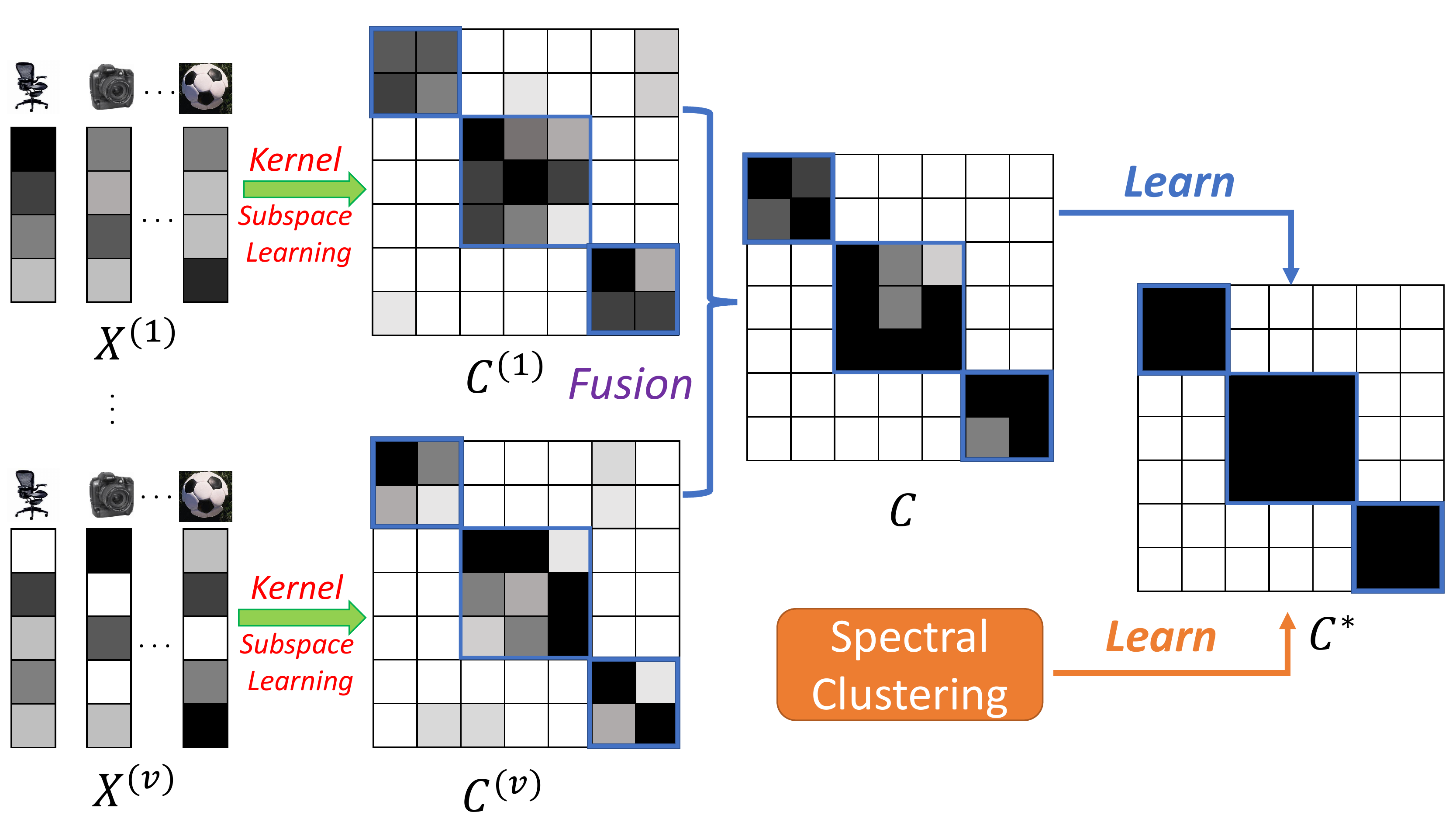}
	\caption{The framework of our proposed method.}
	\label{fig:flow}
\end{figure*}

\section{Multi-View Kernel Subspace Clustering}
\label{sec:subspace}
In this section, we first provide a brief background on sparse subspace
clustering. After that, we will give
the motivation of our proposed method.

\subsection{Sparse Subspace Clustering}
Given $n$ data points $\mX=\{\vx_1,\vx_2,\dots,\vx_n\}\in\Real^{d\times n}$, 
subspace clustering  assumes that
each data point can be approximated by a linear combination of dataset samples \cite{elhamifar2013sparse}:
\begin{equation}
	\mX = \mX\mC+\mE,
\end{equation}
where $\mC=\{\vc_1,\vc_2,\dots,\vc_n\}\in\Real^{n\times n}$  is the subspace representation matrix, with each $\vc_i$
 representing  the
original data point $\x_i$ based on the subspace. $\mE \in \Real^{d\times n}$ is
the error matrix.

Sparse subspace clustering formulates the objective as:
\begin{equation}\label{eq:sc}
	\min_{\mC} \|\mX - \mX\mC\|_F^2+\theta \|\mC\|_1, \ \  \st \ \ diag(\mC)=0, \ \ \mC^T\ones=\ones,
\end{equation}
where $\|\cdot\|_F$ denotes the Frobenius norm while $\|\mC\|_1=\sum_{i,j}|\mC_{ij}|$.  The constraint $\mC^T\ones=\ones$ indicates that the data point lies in a
union of affine subspaces while the
constraint $diag(\mC)=0$ rules out the case that a data point is
represented by itself, which hints that each
data point $\x_i$ can only be represented by a combination of
other points $\x_j (j \ne i)$. Solving the optimization problem in 
Eq. (\ref{eq:sc}), we will get the representation $\vc_i$ for each data point $\x_i$.

After obtaining the subspace structure, we construct the
affinity matrix by setting $\mW =\frac{|\mC|+|\mC|^T}{2}$. Therefore, we
can perform spectral clustering on subspace affinity
matrix:
\begin{equation}\label{eq:spec}
\min_{\mF}\optr{\mF^T\mL\mF}, \quad \st \quad \mF^T\mF=\mId,
\end{equation}
where $\mF$ is the cluster indicator matrix, $\mL:=\mD-\mW$
where $\mD$ is a diagonal matrix given by $\mD(i,i) = \sum_j \mW(i,j)$.

\subsection{Robust Multi-View Kernel Subspace Clustering}
Given the  $v$-view dataset $\mX^{(v)}\in\Real^{d_v\times n}$, if we perform the
subspace learning on each single view, we can get the subspace representation $\mC^{(v)}$ for the $v$-th view. The fundamental challenge boils down to combine multi-view features in
subspace clustering. An intuitive and naive method is to
concatenate all the features together and perform clustering
on the concatenated features,
where the more informative view and the less informative one will
be treated equally. Therefore, the solution is inevitably not optimal in many scenarios. In contrast, 
one can perform the clustering on each single view
 followed by fusing them together. In order to combine multi-view sparse subspace clustering results, we can perform the subspace learning on different views simultaneously by solving:
\begin{equation}\label{eq:mvsc}
	\begin{aligned}
&\min_{\mC^{(v)},\mC^*} \sum_v\|\mX^{(v)} - \mX^{(v)}\mC^{(v)}\|_F^2+\theta \|\mC^{(v)}\|_1\\&+\lambda\|\mC^{(v)}-\mC^*\|^2_F, \ \st \ \  diag(\mC^{(v)})=0, \ \ \mC^{(v)T}\ones=\ones, 
	\end{aligned}
\end{equation}
where $\mC^*$ is the  consensus affinity matrix across
multiple views and spectral clustering will be performed based on it. 
However, experiments have demonstrated that $\mC^{(v)}$ can be significantly different. One can see that if a certain view $\mC^{(v)}$  is not learned well, to minimize the objective, $\mC^{*}$ will deviate from optimal solution due to the squared Frobenius norm which is known to be sensitive to noise/outliers. Inspired by the  observation above, to mitigate its adverse influence, we propose a more robust~\cite{brand2020learning,yang2019learning,liu2018learning} formulation: 
\begin{equation}\label{eq:rmvsc}
\begin{aligned}
	&\min_{\mC^{(v)},\mC^*} \sum_v\|\mX^{(v)} - \mX^{(v)}\mC^{(v)}\|_F^2+\theta \|\mC^{(v)}\|_1\\&+\lambda\|\mC^{(v)}-\mC^*\|_1, \ \st \ \  diag(\mC^{(v)})=0, \ \ \mC^{(v)T}\ones=\ones. 
\end{aligned}
\end{equation}
On the other hand, kernel tricks have been played in various machine learning techniques/algorithms such as  PCA~\cite{mika1998kernel}, SVM~\cite{soman2009machine}, $K$-means~\cite{dhillon2004kernel}, \etc Those kernel version methods yield very promising results especially when the data in original space  is not well separable, but can be separated by projecting into higher dimension space via $\Phi(\cdot)$ where $\Phi: \Real^d \rightarrow\Real^m (m>d)$~\cite{friedman2017elements}. Therefore, we introduce the robust multi-view kernel subspace clustering by optimizing:
\begin{equation}\label{eq:rkmvsc}
	\begin{aligned}
		&\min_{\mC^{(v)},\mC^*} \sum_v\|\Phi(\mX^{(v)}) - \Phi(\mX^{(v)})\mC^{(v)}\|_F^2+\theta \|\mC^{(v)}\|_1\\&+\lambda\|\mC^{(v)}-\mC^*\|_1, \st \ \  diag(\mC^{(v)})=0, \ \ \mC^{(v)T}\ones=\ones. 
	\end{aligned}
\end{equation}
\subsection{Enriched Multi-View Subspace Clustering}
Most existing multi-view subspace learning will do spectral clustering after obtaining $\mC^*$ which ignores the potential connection between the two stages. As a contribution of this paper, we propose an enriched procedure by combining the learning  with clustering stage via:
\begin{equation}\label{eq:erkmvsc}
	\begin{aligned}
	&\min_{\mC^{(v)},\mC^*,\mF} \sum_v\|\Phi(\mX^{(v)}) - \Phi(\mX^{(v)})\mC^{(v)}\|_F^2+\theta \|\mC^{(v)}\|_1\\&+\lambda\|\mC^{(v)}-\mC^*\|_1+\gamma\optr{\mF^T\mL\mF},\\ & \st \quad \mF^T\mF=\mId, \ \  diag(\mC^{(v)})=0, \ \ \mC^{(v)T}\ones=\ones. 	
	\end{aligned}
\end{equation}
It is worth noting that here $\mL$ is constructed based on the affinity matrix $\mW$ from $\mC^*$ instead of $\mC^{(v)}$ in each view. Therefore, different from Eq. (\ref{eq:rkmvsc}) which learns $\mC^*$ from each view, Eq. (\ref{eq:erkmvsc}) also learns $\mC^*$ from spectral clustering.

The proposed optimization model consists of two parts. The
first part is the intra-view structure learning, which aims to learn
the subspace structure in each view. The second part is the inter-view consistency learning, which measures the correlation across
different views. By exploring both the view-specific property and
view-consistency across multi-view data, our unified model can
learn both the intra-view subspace structure and common cluster structure simultaneously. In this way, the proposed method
can achieve the optimal consensus affinity matrix across multiple
views that produces promising clustering results. Fig. \ref{fig:flow} shows the
main framework of the proposed method.
\section{Optimization}
\label{sec:opt}
Considering the constraints and non-differential property of the above objective for $\mC^{(v)}$, we propose an updating algorithm based on Alternating Direction Method of Multipliers (ADMM)~\cite{boyd2011distributed,liu2019learning,deng2016global}.
By introducing $\mA^{(v)}=\mC^{(v)}\in\Real^{n\times n}$, we reformulate the objective as:
\begin{equation}\label{eq:erkmvsc_a}
	\begin{aligned}
		&\min_{\mA^{(v)},\mC^{(v)},\mC^*,\mF} \sum_v\|\Phi(\mX^{(v)}) - \Phi(\mX^{(v)})\mC^{(v)}\|_F^2+\theta \|\mA^{(v)}\|_1\\&+\lambda\|\mA^{(v)}-\mC^*\|_1+\gamma\optr{\mF^T\mL\mF},\\ & \st \quad \mF^T\mF=\mId, diag(\mC^{(v)})=0, \mC^{(v)T}\ones=\ones, \mA^{(v)}=\mC^{(v)}. 	
	\end{aligned}
\end{equation}
The corresponding augmented Lagrangian function is:
\begin{equation}\label{eq:erkmvsc_admm}
	\begin{aligned}
		&\mathcal{L}(\mA^{(v)},\mC^{(v)},\mC^*,\mF,\vdelta^{(v)},\mSigma^{(v)})\\=& \sum_v\|\Phi(\mX^{(v)}) - \Phi(\mX^{(v)})\mC^{(v)}\|_F^2+\lambda\|\mA^{(v)}-\mC^*\|_1\\+&\frac{\rho}{2}\|\mC^{(v)T}\ones-\ones\|^2_2+\lg\vdelta^{(v)},C^{(v)T}\ones-\ones\rg+\theta \|\mA^{(v)}\|_1\\+&\frac{\rho}{2}\|\mC^{(v)}-\mA^{(v)}+diag(\mA^{(v)})\|^2_F+\gamma\optr{\mF^T\mL\mF}\\ +&\lg\mSigma^{(v)},\mC^{(v)}-\mA^{(v)}+diag(\mA^{(v)})\rg,\quad \st \quad \mF^T\mF=\mId, 	
	\end{aligned}
\end{equation}
where $\vdelta\in\Real^n, \mSigma\in\Real^{n\times n}$ are the \textit{Lagrangian Multipliers}.
\subsection{Updating $\mF$}
When fixing $\mC^*$, $\mL$ is fixed and $\mF$ can be optimized via:
\begin{equation}\label{eq:F}
		\min_{\mF} \optr{\mF^T\mL\mF},\quad \st \quad \mF^T\mF=\mId, 	
\end{equation}
where $\mL=\mD-\mW$ and $\mW=\frac{|\mC^*|+|\mC^*|^T}{2}$. Apparently the solutions are the eigenvectors corresponding to the smallest $k$ eigenvalues of the Laplacian matrix $\mL$ where $k$ is the number of clusters~\cite{liu2019spherical}. 
\subsection{Updating $\mA^{(v)}$}
For optimized $\mA$ in each view, it is obtained via:
\begin{equation}\label{eq:A}
	\min_{\mJ} \beta\|\mJ\|_1+\alpha\|\mJ-\mC^*\|_1+\frac{1}{2}\|\mJ-\mY\|_F^2, 
\end{equation}
with $\mA=\mJ-diag(\mJ)$, where $\mY=\mC+\frac{\mSigma}{\rho}, \beta=\frac{\theta}{\rho}, \alpha=\frac{\lambda}{\rho}$.
Apparently, $diag(\mA)=0$ and the above equation can be solved through element-wise optimization:
\begin{equation}\label{eq:a}
	\min_{j} \beta|j|+\alpha|j-c^*|+\frac{1}{2}(j-y)^2. 
\end{equation}
Due to space limit, we provide the closed solution for $c^*\ne0$ (otherwise, it degenerates into well known standard soft-thresholding with $j^*=sgn(y)max\{|y|-\alpha-\beta,0\}$) by leaving the details to supplemental file:
\begin{equation}\label{eq:jsolution}
	j^*=\begin{cases}
		y-\alpha-\beta, \ if \ c^*>0 \wedge y\ge \alpha+\beta+c^*;\\
		y+\alpha-\beta, \ if \ c^*>0 \wedge 0< y+ \alpha-\beta < c^*;\\
		y+\alpha+\beta, \ if \ c^*>0 \wedge 0\ge y+ \alpha+\beta;\\
		y-\alpha-\beta, \ if \ c^*< 0 \wedge y\ge \alpha+\beta;\\
		y-\alpha+\beta, \ if \ c^*<0 \wedge 0> y-\alpha+\beta > c^*;\\
		y+\alpha+\beta, \ if \ c^*<0 \wedge c^*\ge y+ \alpha+\beta;\\
		0, else,
	\end{cases}
\end{equation}
where `$\wedge$' denotes \textit{logical conjunction}.
\subsection{Updating  $\mC^{(v)}$}
For optimized $\mC$ in each view, it can be obtained via (by skipping $diag(\mA)$  as it is $0$ aforementioned):
\begin{equation}\label{eq:C}\small
	\min_{\mC} \|\Phi(\mX) - \Phi(\mX)\mC\|_F^2+\frac{\rho}{2}\|\mC^T\ones-\ones+\frac{\vdelta}{\rho}\|^2_2+\frac{\rho}{2}\|\mC-\mA+\frac{\mSigma}{\rho}\|^2_F.
\end{equation}
By taking the derivative and set it to be 0, we have \footnote{We note that for Linear Kernel case, which is $\Phi(\mX)=\mX\in\Real^{d\times n}$ and $\mathcal{K}(\mX,\mX)=\mX^T\mX$. When $n\gg d$, we have accelerated updating algorithm for inversion calculation. First we denote $\mX^T\mX+\rho\mId+\rho\ones\ones^T=\mZ^T\mZ+\rho\mId$, where $\mZ=[\mX;\sqrt{\rho}\ones^T]\in\Real^{(d+1)\times n}$. Then by matrix inversion lemma (aka Sherman-Morrison-Woodbury Formula), $(\mZ^T\mZ+\rho\mId_n)^{-1}=\rho^{-1}\mId_n-\rho^{-2}\mZ^T(\mId_{d+1}+\rho\mZ\mZ^T)^{-1}\mZ$, the complexity can be reduced from $\mathcal{O}(n^3)$ to $\mathcal{O}(d^3+dn^2)$ which is a significant improvement for $n\gg d$.}:
\begin{equation}
	\mC=(\mathcal{K}+\rho\mId+\rho\ones\ones^T)^{-1}(\mathcal{K}+\rho\ones\ones^T-\ones\vdelta^T+\rho\mA-\mSigma),
\end{equation}
where $\mathcal{K}=\Phi(\mX)^T\Phi(\mX)$. One can see that with different kernel chosen, $\mathcal{K}$ is different but always are computationally efficient. For example, when polynomial kernel is applied, then $\mathcal{K}(i,j)=(\lg \x_i,\x_j\rg+c)^d$.
\subsection{Updating $\mC^*$}
As $\mC^*$ is enriched, which is related with 2 terms, it can be optimized via:
\begin{equation}\label{eq:c_start_ori}
	\min_{\mC^*} \gamma\optr{\mF^T\mL\mF}+\sum_v \lambda \|\mA^{(v)}-\mC^*\|_1.
\end{equation}
Before we optimize the above equation, we first introduce a useful lemma which is critical for $\mC^*$:
\begin{lem}
	For Laplacian matrix $\mL$ and the matrix $\mF$, we have:
	\begin{equation}
		\optr{\mF^T\mL\mF}=\frac{1}{2}\sum_{i,j}\mW(i,j)\|\vf_i-\vf_j\|^2_2.
	\end{equation}
\end{lem}

\noindent We turn to optimize $\mC^*$ by noticing the above equation can be written in a more compact formulation: $\optr{\mF^T\mL\mF}=\frac{1}{2}\lg \mW,\mQ\rg$, where $\mQ$ is symmetric and $\mQ(i,j)=\|\vf_i-\vf_j\|^2_2$. On the other hand, by definition  $\mW=\frac{|\mC^*|+|\mC^*|^T}{2}$, by simple algebraic operation we have:
\begin{equation}
\optr{\mF^T\mL\mF}=\frac{1}{2}\lg |\mC^*|,\mQ\rg.
\end{equation}
Therefore, $\mC^*$ can be optimized by:
\begin{equation}\label{eq:c_star}
	\min \frac{\gamma}{2}\lg |\mC^*|,\mQ\rg+\sum_v \lambda \|\mA^{(v)}-\mC^*\|_1.
\end{equation}
Similar to $\mA$, we can optimize $\mC^*$ by element-wise:
\begin{equation}\label{c_start}
	\min \gamma q|c^*|+\sum_v 2\lambda |a^{(v)}-c^*|.
\end{equation}
Without loss of generality, we sort $[a^{(1)}, a^{(2)},\dots,a^{(v)}]$  in non-decreasing order as $[a_1, a_2,\dots,a_v]$ and none is zero (or it can be transferred into this case by simple operation). Due to space limit, we leave the derivative details to supplemental file and directly give the solution\footnote{It is worth noting that the optimal solution may not be unique. In practice, one can visit all $a_i (1\le i\le v)$ in addition to $0$, and  simply set $c^*$ as $a_i$ or $0$ which yields the lowest objective in Eq. (\ref{c_start}).}:
\begin{equation}\label{eq:csolution}
	c^*=\begin{cases}
		a_{\left \lceil \frac{2v\lambda-\gamma q}{4\lambda} \right \rceil}, \ if \ 2v\lambda>\gamma q \wedge a_{\left \lceil \frac{2v\lambda-\gamma q}{4\lambda} \right \rceil}>0; \\
		a_{\left \lceil \frac{2v\lambda+\gamma q}{4\lambda} \right \rceil}, \ if \ 2v\lambda>\gamma q \wedge a_{\left \lceil \frac{2v\lambda+\gamma q}{4\lambda} \right \rceil}<0;\\
		0, else,
	\end{cases}
\end{equation}
where $\left \lceil \cdot \right \rceil$ denotes the ceiling function.
\subsection{Updating Lagrangian Multipliers in Each View}
Following ADMM framework~\cite{boyd2011distributed}, we can simply update Lagrangian Multipliers by gradient ascent:
\begin{equation}\label{eq:lm}
	\begin{aligned}
		&\vdelta^{(v)}=\vdelta^{(v)}+\rho(\mC^{(v)T}\ones-\ones),\\
		&\mSigma^{(v)}=\mSigma^{(v)}+\rho(\mC^{(v)}-\mA^{(v)}).
	\end{aligned}
\end{equation}
We summarize the above algorithm in Alg. ~\ref{alg:alg}.
\begin{algorithm}[tb]
	\caption{Algorithm for Enriched Robust Multi-View Kernel Subspace Clustering to solve Eq. (\ref{eq:erkmvsc}).}
	\label{alg:alg}
	\begin{algorithmic}
		\State {\bfseries Input:} data $\mX^{(v)}\in\Real^{d_v\times n}$, number of clusters $k$, regularization parameters $\lambda,\gamma,\theta$, number of iterations $T$.
		\State {\bfseries Initialization:} $\mC^{(v)}, \mSigma^{(v)}, \mA^{(v)}, \mC^* \in\Real^{n\times n}, \vdelta^{(v)}\in\Real^{n}, \rho=0.2, t=1$.
		\While{$t \leq T$}
		\State Optimize $\mF$ by solving Eq.~(\ref{eq:F});
		\State Optimize  $\mA$  in each view by solving Eq.~(\ref{eq:A});
		\State Optimize  $\mC$  in each view by solving Eq.~(\ref{eq:C});
		\State Optimize  each  $\mC^*$  by solving Eq.~(\ref{eq:c_star});
		\State Update $\vdelta, \mSigma$ in each view as Eq.~(\ref{eq:lm});
		\State Update $\rho=1.2\rho$;
		\State $t=t+1$.
		\EndWhile
		\State {\bfseries Output:} $\mF$, based on which $K$-means will be conducted after row normalization.
	\end{algorithmic}
\end{algorithm}
\begin{table*}[t]
	\footnotesize
	\begin{center}
		\caption{Subspace clustering results on various benchmark datasets}
		\label{tab:1}
		\begin{tabular}{*{9}{c}}
			\toprule
			\multicolumn{1}{c}{}  &
			\multicolumn{2}{c}{MSRC-v1}  &
			\multicolumn{2}{c}{Handwritten} &
			\multicolumn{2}{c}{Caltech101-7}  &
			\multicolumn{2}{c}{Caltech101-20}\\
			\cmidrule(r){1-1}\cmidrule(r){2-3}\cmidrule(r){4-5}\cmidrule(r){6-7}\cmidrule(r){8-9}
			{Method} & {ACC} & {NMI} & {ACC} & {NMI} & {ACC} & {NMI} & {ACC} & {NMI}\\
			\midrule
			HOG-SC & 0.597$\pm$0.057 & 0.502$\pm$0.027 & 0.632$\pm$0.071 & 0.518$\pm$0.074 & 0.609$\pm$0.051 & 0.561$\pm$0.039 & 0.307$\pm$0.062 & 0.287$\pm$0.025\\
			CEN-SC & 0.618$\pm$0.038 & 0.556$\pm$0.017 & 0.711$\pm$0.027 & 0.641$\pm$0.041 & 0.657$\pm$0.032 & 0.587$\pm$0.031 & 0.501$\pm$0.018 & 0.536$\pm$0.019\\
			CMT-SC & 0.331$\pm$0.052 & 0.203$\pm$0.056 & 0.213$\pm$0.087 & 0.198$\pm$0.077 & 0.391$\pm$0.097 & 0.281$\pm$0.011 & 0.214$\pm$0.051 & 0.258$\pm$0.028\\
			LBP-SC & 0.587$\pm$0.061 & 0.525$\pm$0.038 & 0.303$\pm$0.043 & 0.321$\pm$0.076 & 0.458$\pm$0.082 & 0.322$\pm$0.059 & 0.301$\pm$0.022 & 0.317$\pm$0.028\\
			GIST-SC & 0.309$\pm$0.049 & 0.281$\pm$0.027 & 0.288$\pm$0.036 & 0.212$\pm$0.081 & 0.402$\pm$0.073 & 0.378$\pm$0.054 & 0.262$\pm$0.057 & 0.209$\pm$0.062\\
			Gabor-SC &  &  & 
			0.328$\pm$0.059 & 0.277$\pm$0.066 & 0.397$\pm$0.071 & 0.306$\pm$0.039 & 0.296$\pm$0.062 & 0.309$\pm$0.039\\
			\midrule
			HOG-LRR & 0.611$\pm$0.015 & 0.572$\pm$0.018 & 0.512$\pm$0.056 & 0.413$\pm$0.072 & 0.622$\pm$0.013 & 0.508$\pm$0.023 & 0.312$\pm$0.011 & 0.257$\pm$0.011\\
			CEN-LRR & 0.457$\pm$0.018 & 0.323$\pm$0.007 & 0.581$\pm$0.019 & 0.512$\pm$0.017 & 0.615$\pm$0.005 & 0.479$\pm$0.012 & 0.467$\pm$0.012 & 0.472$\pm$0.004\\
			CMT-LRR & 0.343$\pm$0.031 & 0.201$\pm$0.009 & 0.182$\pm$0.072 & 0.131$\pm$0.025 & 0.322$\pm$0.005 & 0.282$\pm$0.019 & 0.276$\pm$0.012 & 0.297$\pm$0.005\\
			LBP-LRR & 0.627$\pm$0.012 & 0.477$\pm$0.016 & 0.219$\pm$0.057 & 0.238$\pm$0.052 & 0.423$\pm$0.009 & 0.327$\pm$0.014 & 0.281$\pm$0.005 & 0.315$\pm$0.009\\
			GIST-LRR & 0.318$\pm$0.021 & 0.196$\pm$0.011 & 0.256$\pm$0.061 & 0.291$\pm$0.015 & 0.399$\pm$0.012 & 0.301$\pm$0.009 & 0.269$\pm$0.011 & 0.291$\pm$0.007\\
			Gabor-LRR &  &  & 
			0.302$\pm$0.049 & 0.256$\pm$0.068 & 0.307$\pm$0.019 & 0.217$\pm$0.011 & 0.302$\pm$0.011 & 0.272$\pm$0.013\\
			\midrule
			P-CoReg & 0.781$\pm$0.008 & 0.691$\pm$0.015 & 0.767$\pm$0.017 & 0.711$\pm$0.039 & 0.678$\pm$0.031 & 0.677$\pm$0.022 & 0.551$\pm$0.021 & 0.601$\pm$0.015\\
			C-CoReg & 0.767$\pm$0.006 & 0.678$\pm$0.031 & 0.758$\pm$0.015 & 0.704$\pm$0.041 & 0.658$\pm$0.052 & \textbf{0.687$\pm$0.017} & 0.502$\pm$0.029 & 0.558$\pm$0.021\\
			RMSC & 0.771$\pm$0.012 & 0.652$\pm$0.017 & 0.775$\pm$0.009 & 0.714$\pm$0.012 & 0.667$\pm$0.019 & 0.676$\pm$0.029 & 0.451$\pm$0.018 & 0.389$\pm$0.017\\
			MCGC & 0.682$\pm$0.012 & 0.601$\pm$0.009 & 0.768$\pm$0.011 & 0.721$\pm$0.021 & 0.649$\pm$0.035 & 0.533$\pm$0.071 & 0.425$\pm$0.033 & 0.392$\pm$0.019\\
			MNMF & 0.657$\pm$0.007 & 0.597$\pm$0.017 & 0.689$\pm$0.052 & 0.518$\pm$0.039 & 0.652$\pm$0.027 & 0.502$\pm$0.019 & 0.455$\pm$0.012 & 0.388$\pm$0.021\\
			Our & \textbf{0.822$\pm$0.037} & \textbf{0.712$\pm$0.022} & \textbf{0.831$\pm$0.028} & \textbf{0.798$\pm$0.037} & \textbf{0.693$\pm$0.057} & 0.671$\pm$0.038 & \textbf{0.576$\pm$0.025} & \textbf{0.657$\pm$0.022}\\
			\bottomrule
		\end{tabular}
	\end{center}
\end{table*}

\begin{table}
	\footnotesize
	\begin{center}
		\caption{Subspace clustering results on benchmark datasets}
		\label{tab:2}
		\begin{tabular}{*{5}{c}}
			\toprule
			\multicolumn{1}{c}{}  &
			\multicolumn{2}{c}{Yale}  &
			\multicolumn{2}{c}{ORL}\\
			\cmidrule(r){1-1}\cmidrule(r){2-3}\cmidrule(r){4-5}
			{Method} & {ACC} & {NMI} & {ACC} & {NMI}\\
			\midrule
			IT-SC & 0.271$\pm$0.026 & 0.256$\pm$0.031 & 0.353$\pm$0.012 & 0.559$\pm$0.015\\
			LBP-SC & 0.617$\pm$0.036 & 0.637$\pm$0.017 & 0.713$\pm$0.009 & 0.798$\pm$0.011\\
			Gabor-SC & 0.653$\pm$0.012 & 0.643$\pm$0.011 & 0.647$\pm$0.011 & 0.813$\pm$0.006\\
			\midrule
			IT-LRR & 0.153$\pm$0.022 & 0.101$\pm$0.007 & 0.108$\pm$0.013 & 0.138$\pm$0.017\\
			LBP-LRR & 0.631$\pm$0.019 & 0.602$\pm$0.022 & 0.722$\pm$0.018 & 0.855$\pm$0.015\\
			Gabor-LRR & 0.625$\pm$0.023 & 0.637$\pm$0.018 & 0.721$\pm$0.021 & 0.823$\pm$0.012\\
			\midrule
			P-CoReg & 0.635$\pm$0.015 & 0.667$\pm$0.037 & 0.723$\pm$0.005 & \textbf{0.868$\pm$0.012}\\
			C-CoReg & 0.655$\pm$0.009 & 0.637$\pm$0.011 & 0.731$\pm$0.007 & 0.852$\pm$0.008\\
			RMSC & 0.630$\pm$0.012 & 0.644$\pm$0.014 & 0.725$\pm$0.015 & 0.825$\pm$0.011\\
			MCGC & 0.649$\pm$0.035 & 0.533$\pm$0.071 & 0.425$\pm$0.033 & 0.392$\pm$0.019\\
			MNMF & 0.564$\pm$0.031 & 0.571$\pm$0.027 & 0.625$\pm$0.013 & 0.798$\pm$0.008\\
			Our & \textbf{0.668$\pm$0.012} & \textbf{0.672$\pm$0.009} & \textbf{0.732$\pm$0.025} & 0.863$\pm$0.019\\
			\bottomrule
		\end{tabular}
	\end{center}
\end{table}
\section{Experiments}
\label{exp}
In this section, we will evaluate our proposed algorithm on several
widely used benchmark datasets to illustrate its potential in multi-view clustering.

Six benchmark datasets are used in the experiment, including MSRC-v1, UCI Handwritten digits \cite{Dua:2019}, Caltech101-7 \cite{fei2006one}, Caltech101-20 \cite{fei2006one}, ORL \cite{samariaproceedings} and Yale \cite{yale}. For each dataset, multiple feature sets are available to describe the images from various aspects. The detailed information is summarized in Table~\ref{tab:dataset}.

Throughout the experiments, we use Matlab R2019a on a laptop with 1.4 GHz QuadCore Intel Core i5 processor. 
The clustering quality is measured by clustering accuracy (ACC), which is the percentage of items correctly clustered with the maximum bipartite matching~\cite{xu2003document}, and normalized mutual information (NMI)~\cite{lancichinetti2009detecting,liu2018high}. We repeat each experiment 10 times and report the average performance with the standard deviation. 
\begin{table*}[ht]
	\begin{center}
		\caption{Datasets information and available feature sets}
		\label{tab:dataset}
		\begin{tabular}{*{10}{c}}
			\toprule
			\multicolumn{1}{c}{Dataset}  &
			\multicolumn{1}{c}{\# images}  &
			\multicolumn{1}{c}{\# classes} &
			\multicolumn{1}{c}{HOG} &
			\multicolumn{1}{c}{CENTRIST} &
			\multicolumn{1}{c}{Color Moment}  &
			\multicolumn{1}{c}{LBP} &
			\multicolumn{1}{c}{GIST} &
			\multicolumn{1}{c}{Intensity} &
			\multicolumn{1}{c}{Gabor} \\
			\midrule
			MSRC-v1 & 210 & 7 &  \checkmark & \checkmark & \checkmark & \checkmark & \checkmark &  & \\
			Handwritten & 2000 & 10 & \checkmark & \checkmark & \checkmark & \checkmark & \checkmark & & \checkmark\\
			Caltech101-7 & 1474 & 7 & \checkmark & \checkmark & \checkmark & \checkmark & \checkmark & & \checkmark\\
			Caltech101-20 & 2386 & 20 & \checkmark & \checkmark & \checkmark & \checkmark & \checkmark &  & \checkmark\\
			ORL & 400 & 40 &  & &  & \checkmark &  & \checkmark & \checkmark\\
			Yale & 165 & 15 &  &  &  & \checkmark &  & \checkmark & \checkmark\\
			\bottomrule
		\end{tabular}
	\end{center}
\end{table*}

%
%

\begin{figure*}[ht]
	\centering 
	\begin{subfigure}{0.25\textwidth}
		\includegraphics[width=1.1\linewidth]{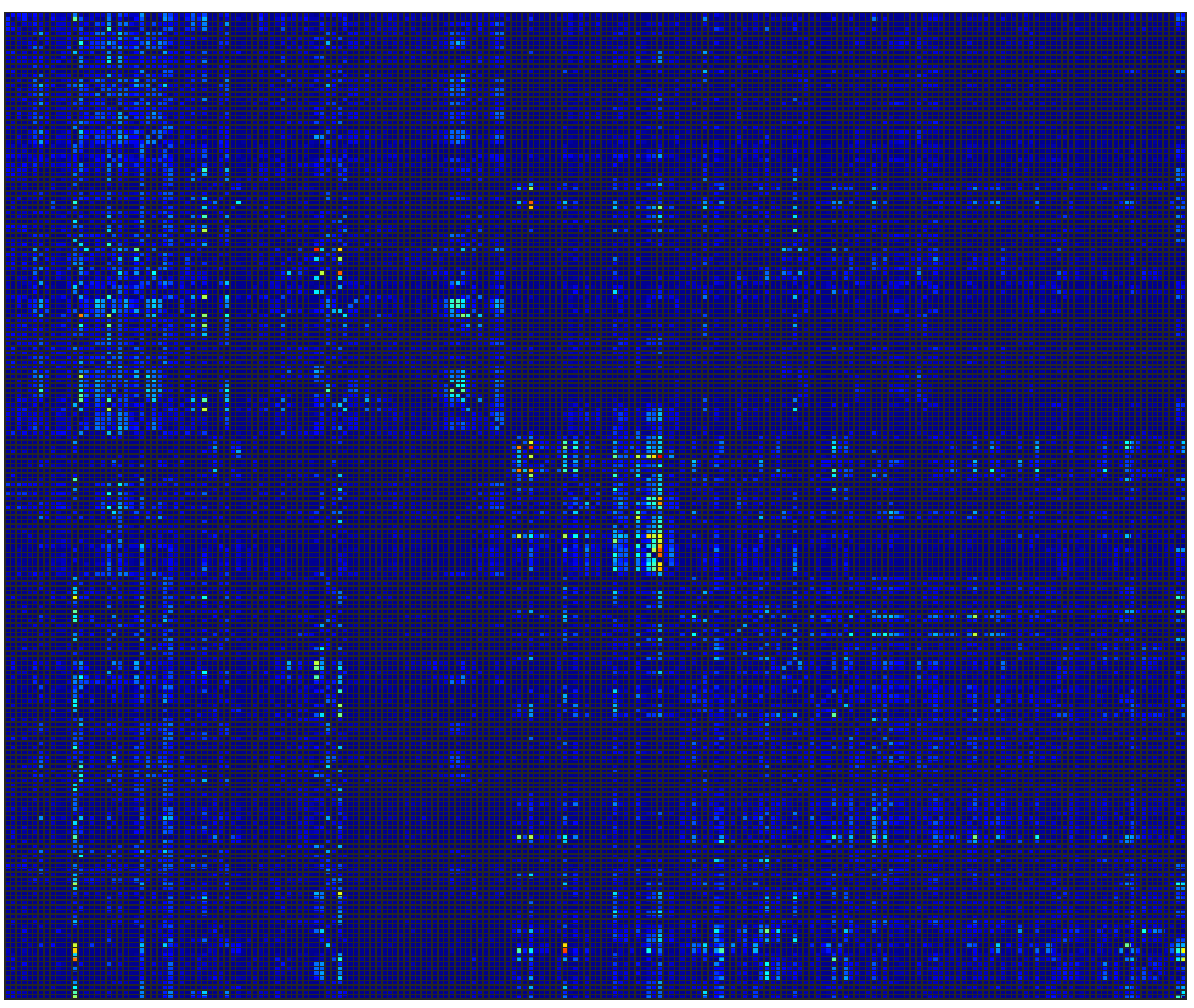}
		\caption{Color moment}
		\label{fig:1}
	\end{subfigure}\hfil 
	\begin{subfigure}{0.25\textwidth}
		\includegraphics[width=1.1\linewidth]{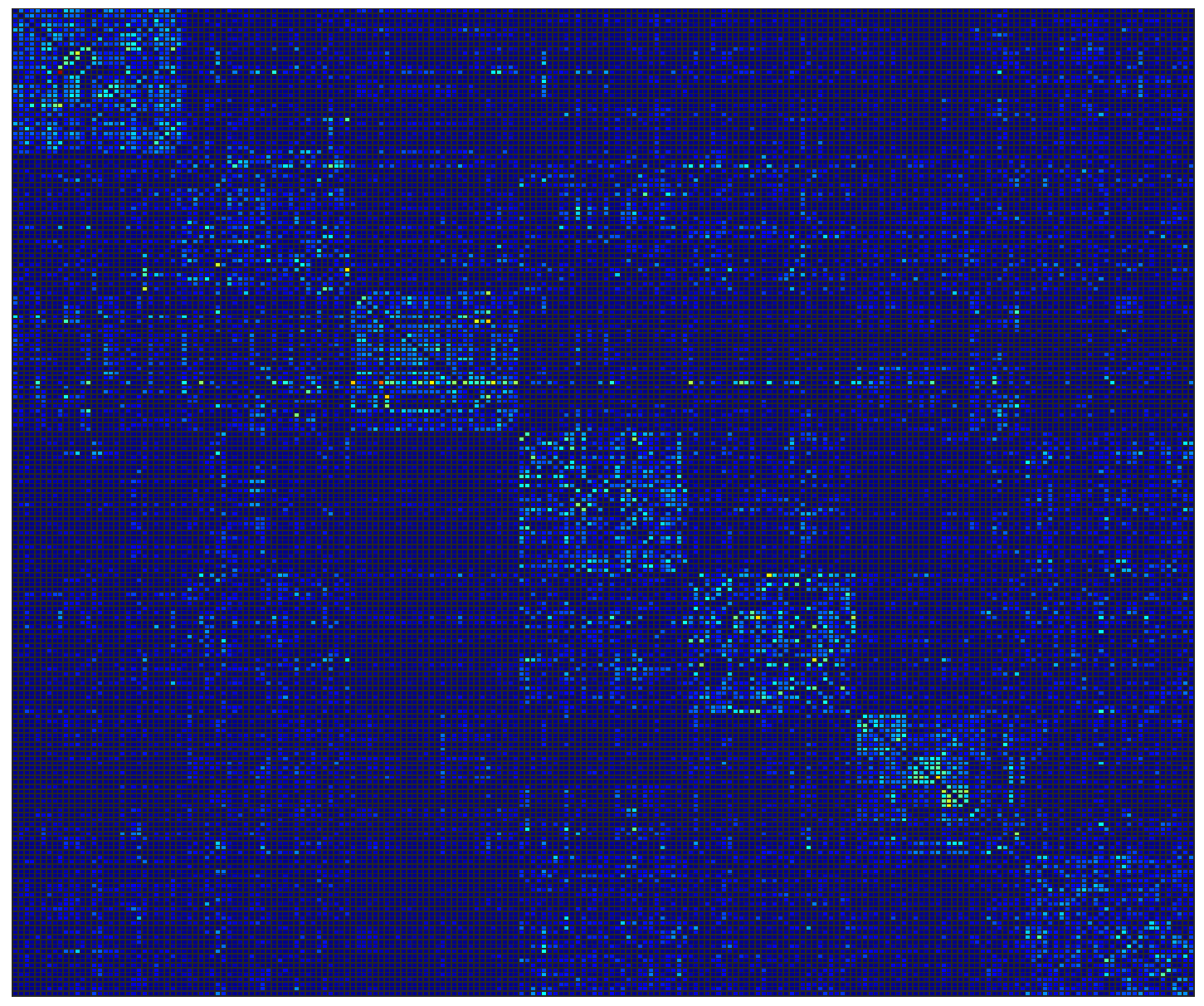}
		\caption{HOG}
		\label{fig:2}
	\end{subfigure}\hfil 
	\begin{subfigure}{0.25\textwidth}
		\includegraphics[width=1.25\linewidth]{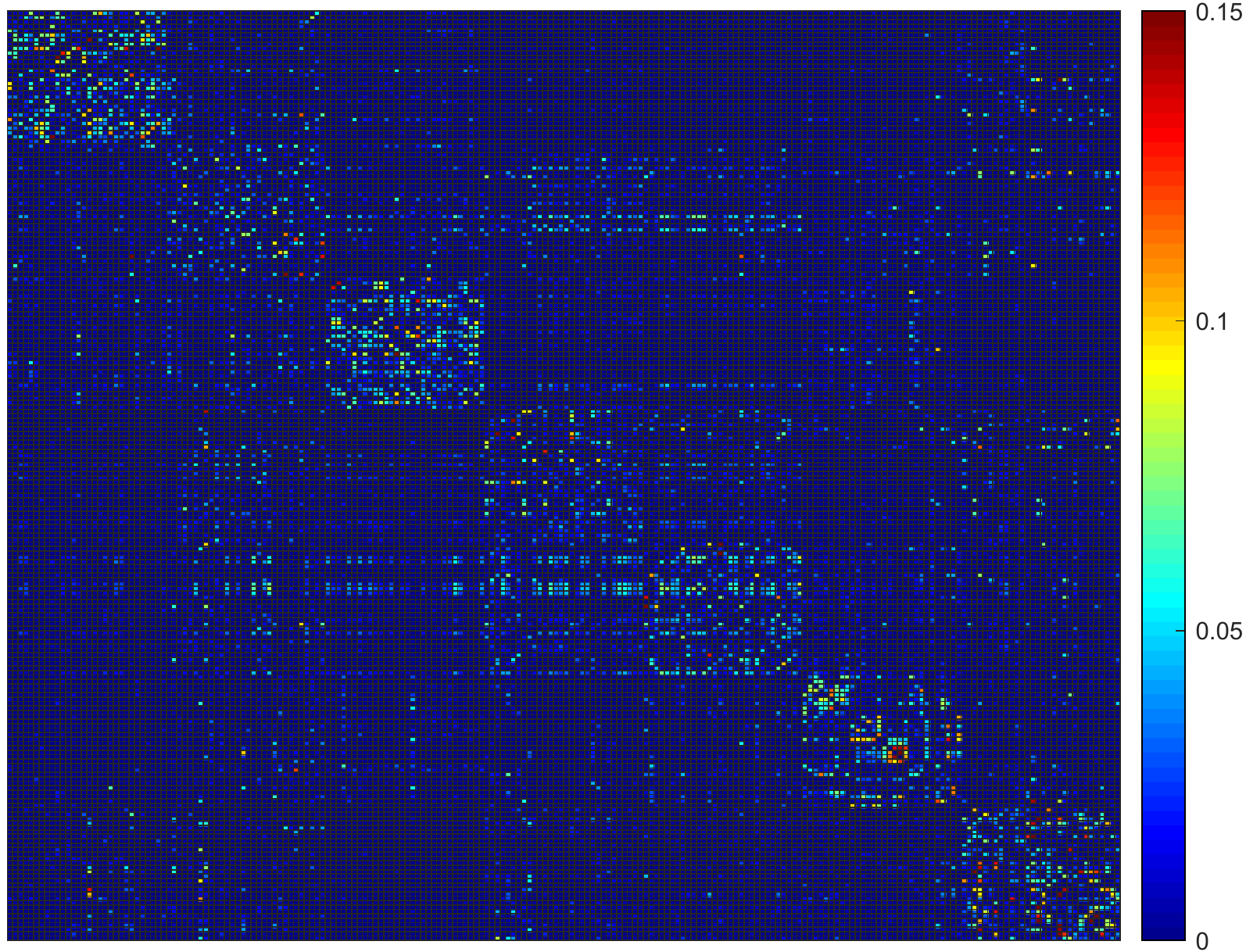}
		\caption{CENTRIST}
		\label{fig:3}
	\end{subfigure}
	
	\medskip
	\begin{subfigure}{0.25\textwidth}
		\includegraphics[width=1.1\linewidth]{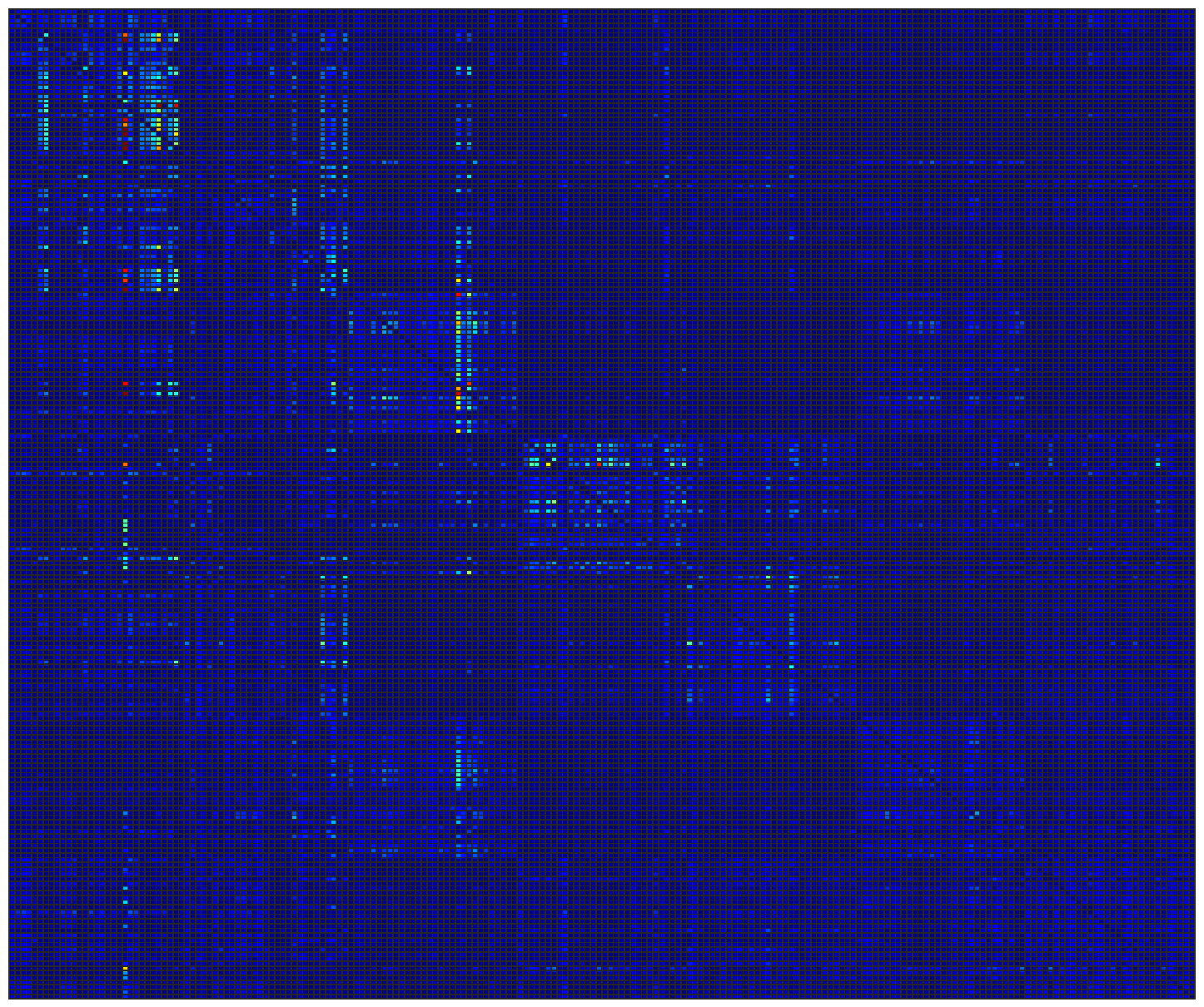}
		\caption{LBP}
		\label{fig:4}
	\end{subfigure}\hfil 
	\begin{subfigure}{0.25\textwidth}
		\includegraphics[width=1.1\linewidth]{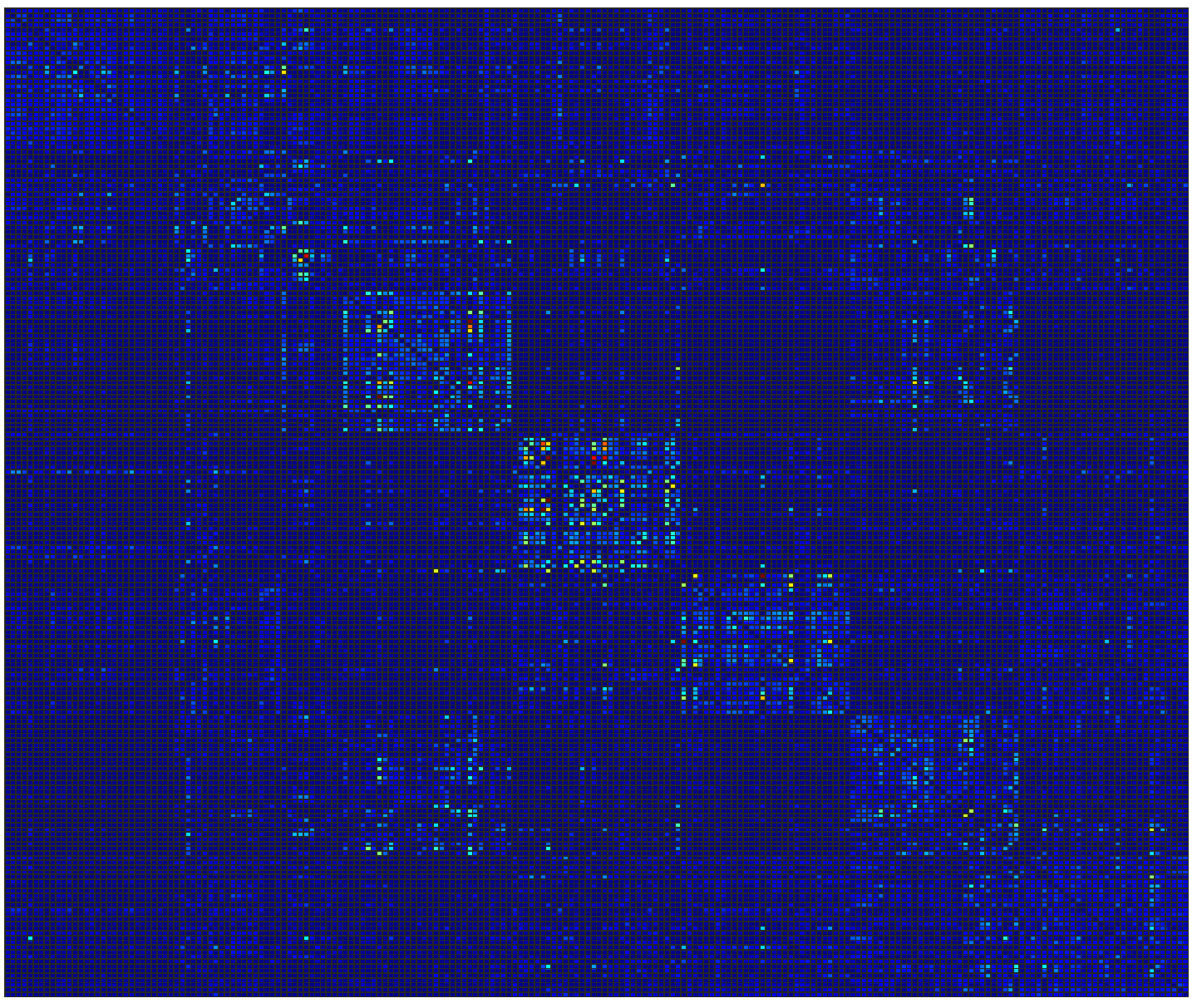}
		\caption{GIST}
		\label{fig:5}
	\end{subfigure}\hfil 
	\begin{subfigure}{0.25\textwidth}
		\includegraphics[width=1.25\linewidth]{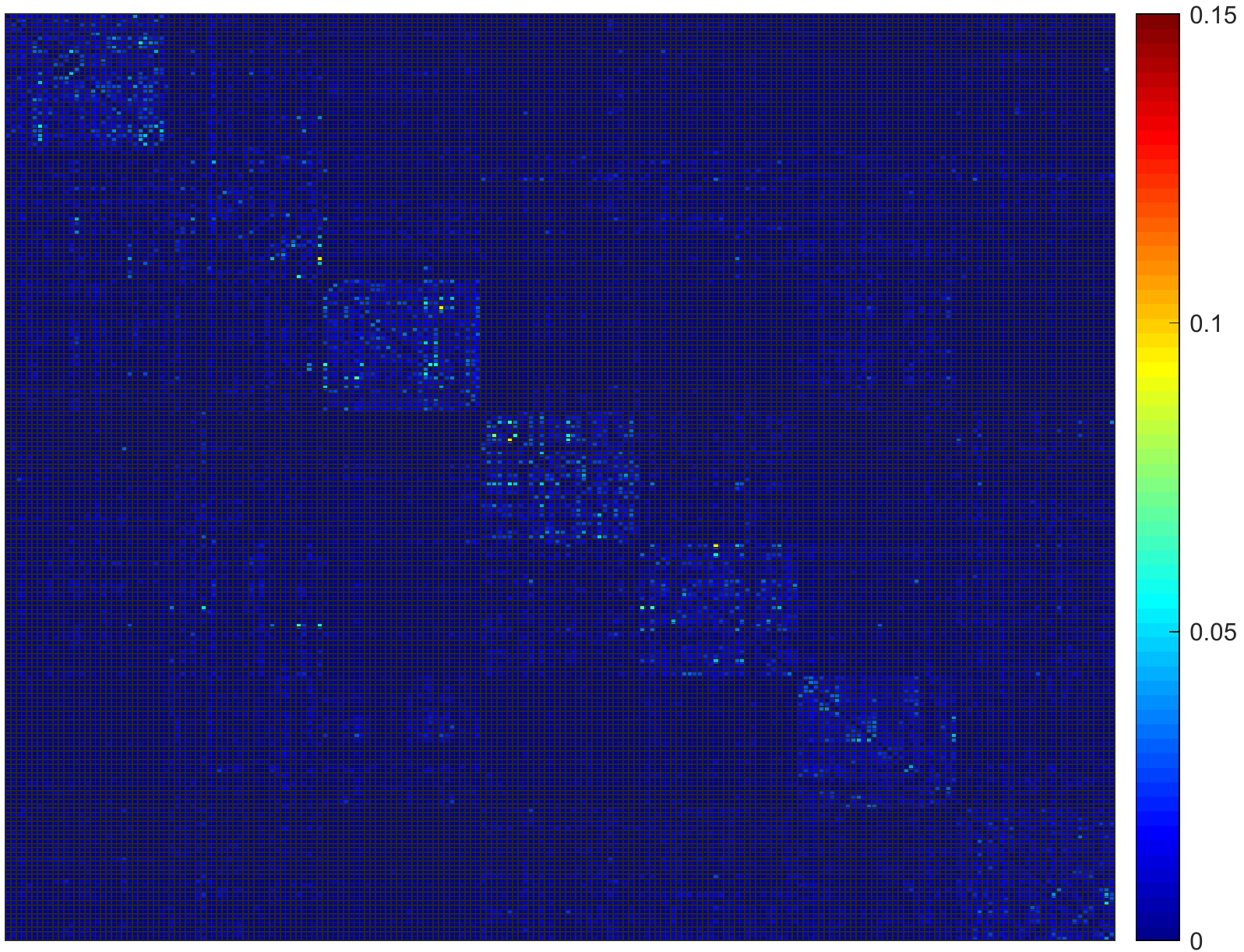}
		\caption{Consensus}
		\label{fig:6}
	\end{subfigure}
	\caption{View-specific and consensus subspace representation matrices on MSRC-v1 dataset. (\textcolor{red}{\textbf{Please zoom in to observe.}})}
	\label{fig:C}
\end{figure*}
\begin{figure*}[ht]
	\centering 
	\begin{subfigure}{0.32\textwidth}
		\includegraphics[width=.95\linewidth]{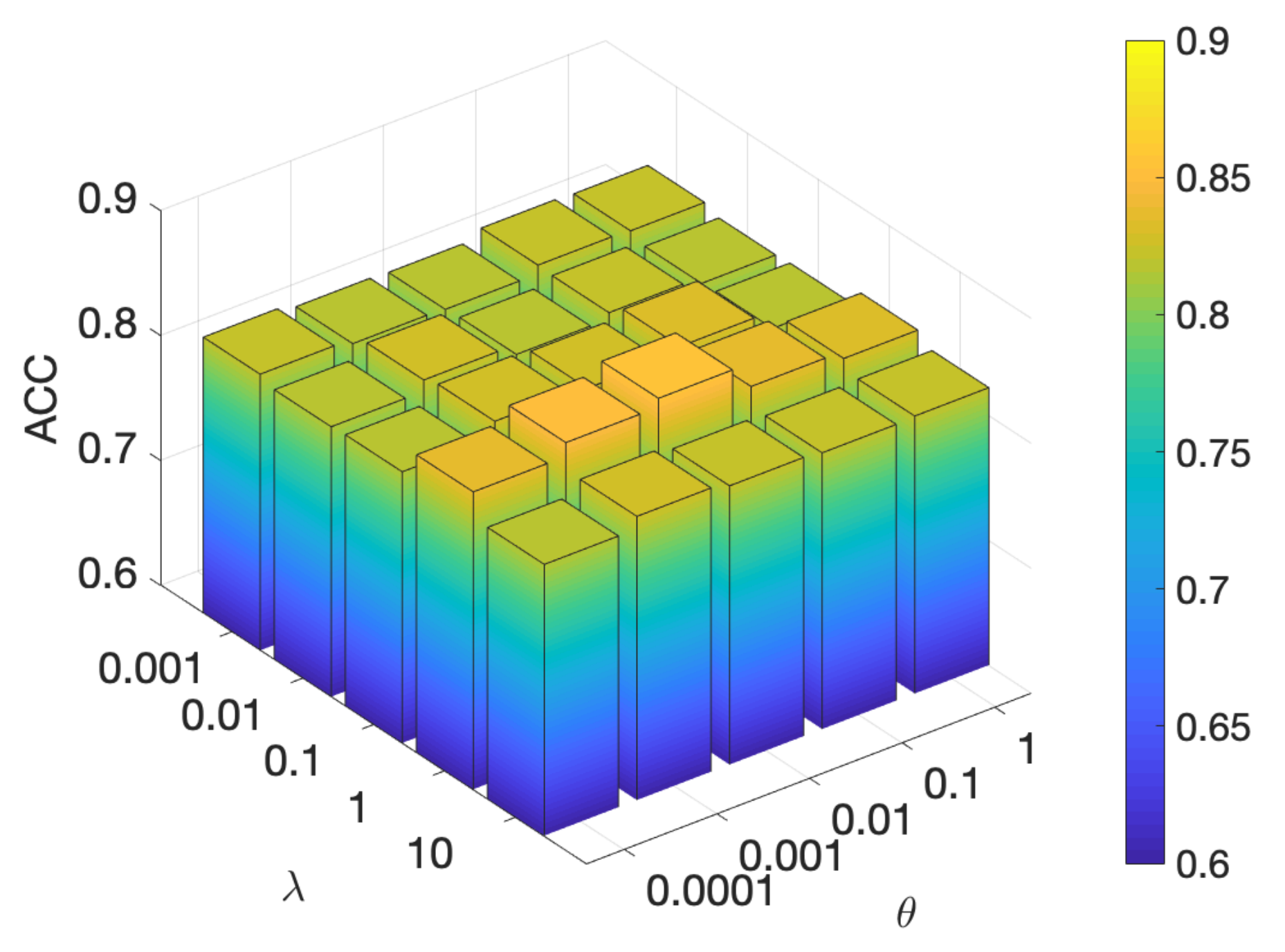}
		\caption{Fixing $\gamma$ to 0.01}
		\label{fig:11}
	\end{subfigure}\hfil 
	\begin{subfigure}{0.32\textwidth}
		\includegraphics[width=.95\linewidth]{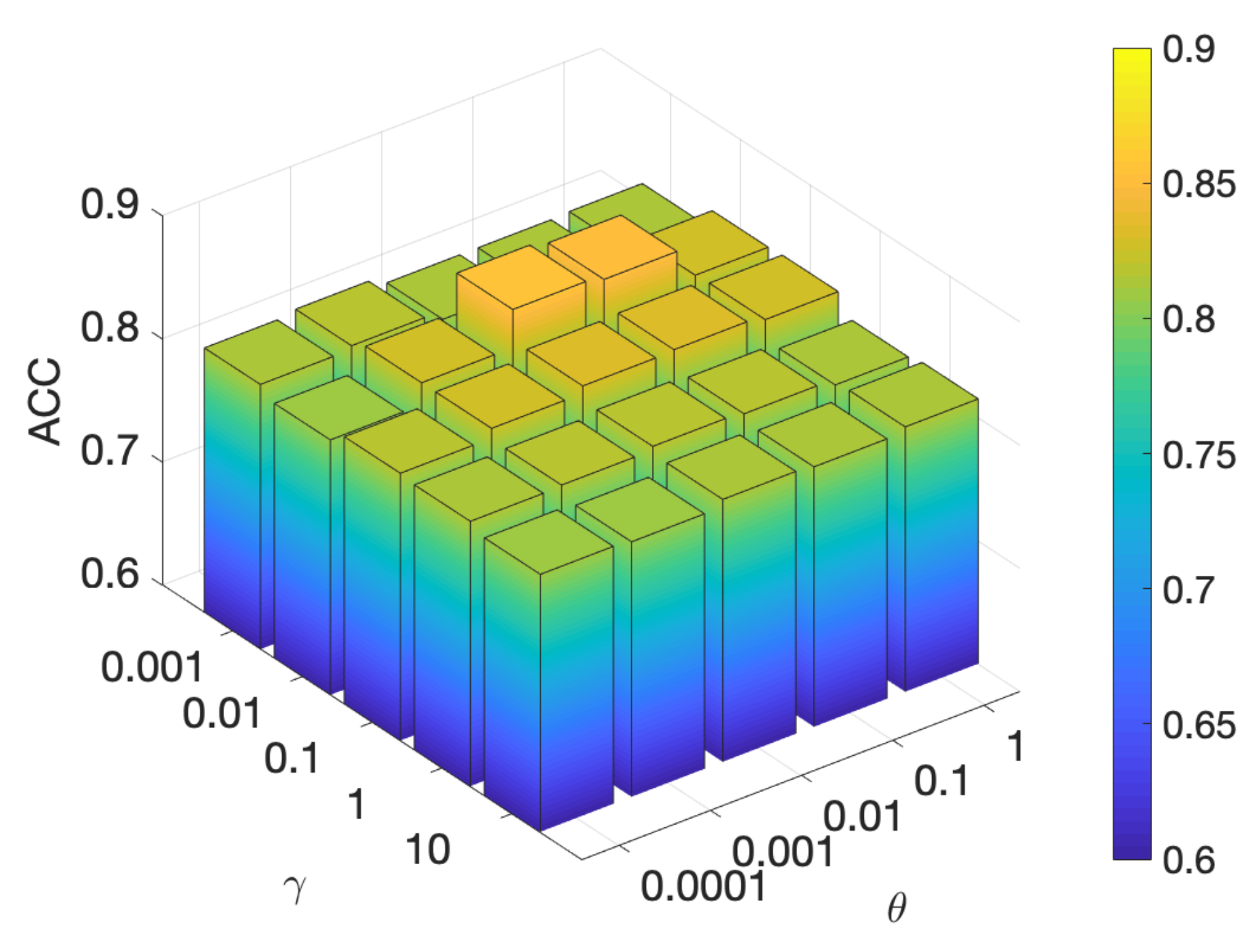}
		\caption{Fixing $\lambda$ to 1}
		\label{fig:12}
	\end{subfigure}\hfil 
	\begin{subfigure}{0.32\textwidth}
		\includegraphics[width=1.1\linewidth]{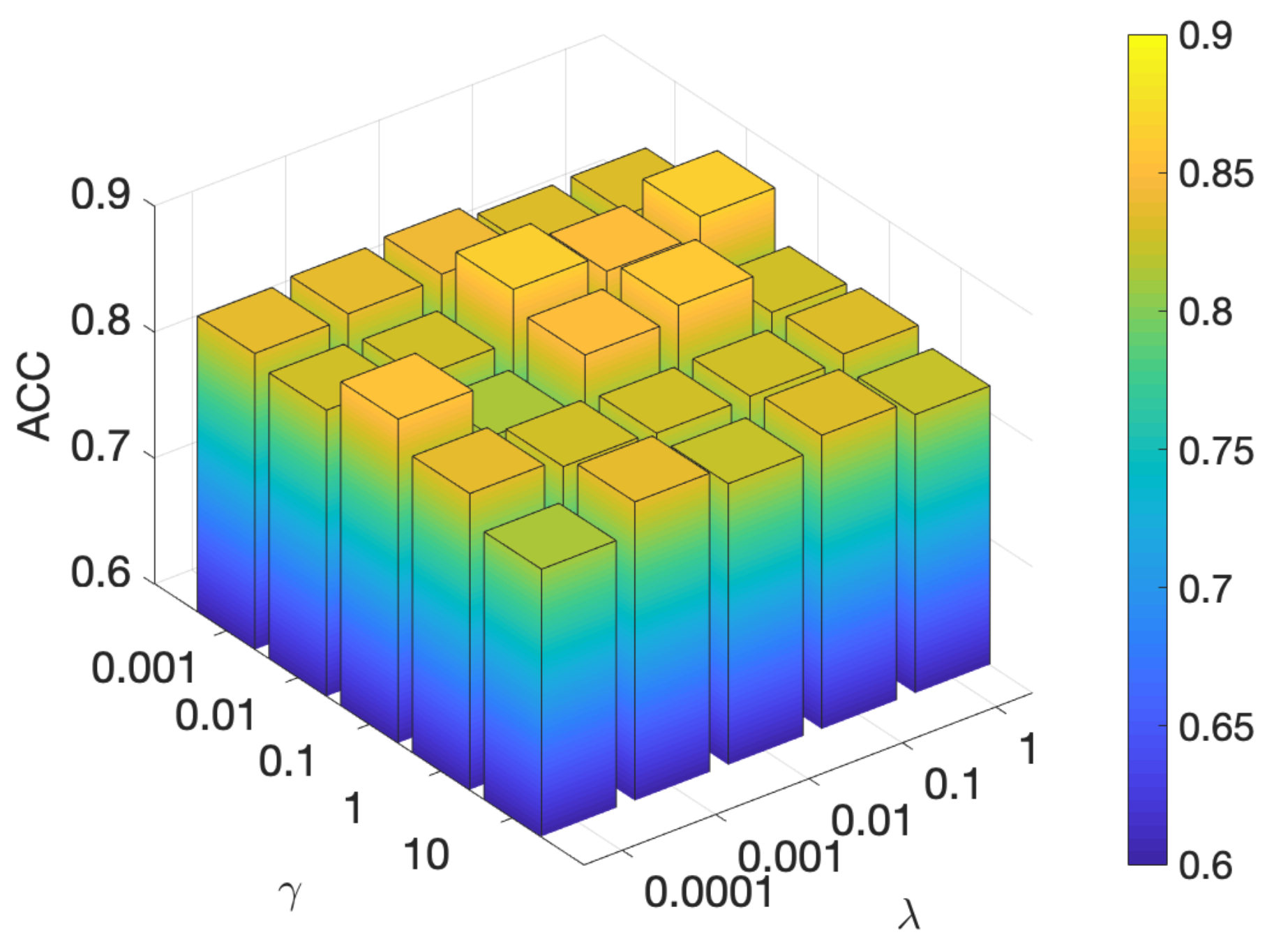}
		\caption{Fixing $\theta$ to 0.01}
		\label{fig:13}
	\end{subfigure}
	\caption{Ablation study -- the influence of regularization parameters on clustering accuracy.}
	\label{fig:cubic}
\end{figure*}
\begin{figure*}[ht]
	\centering
	\begin{minipage}[t]{0.68\textwidth}
		\centering
		\includegraphics[height=4.3cm]{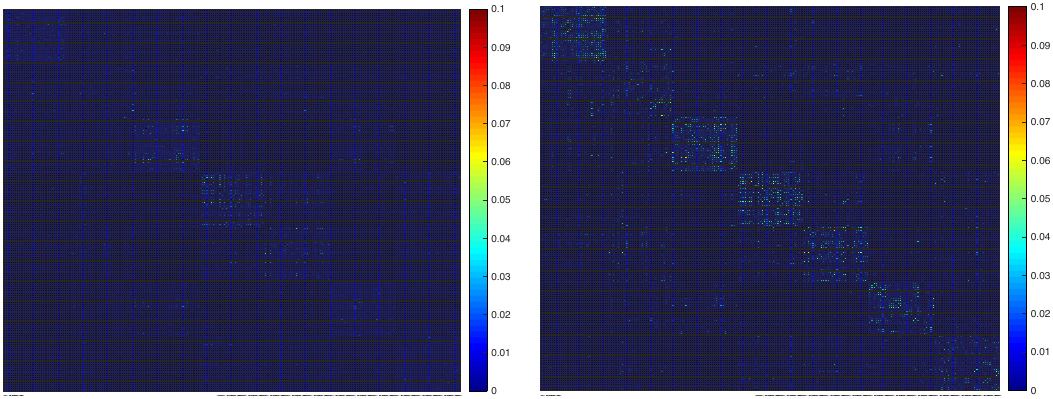}
		\caption{$\mC^*$ obtained via Linear Kernel and Polynomial Kernel. (\textcolor{red}{\textbf{Please zoom in to observe.}})}
		\label{fig:plot}
	\end{minipage}
	\begin{minipage}[t]{0.3\textwidth}
		\centering
		\includegraphics[height=4.3cm]{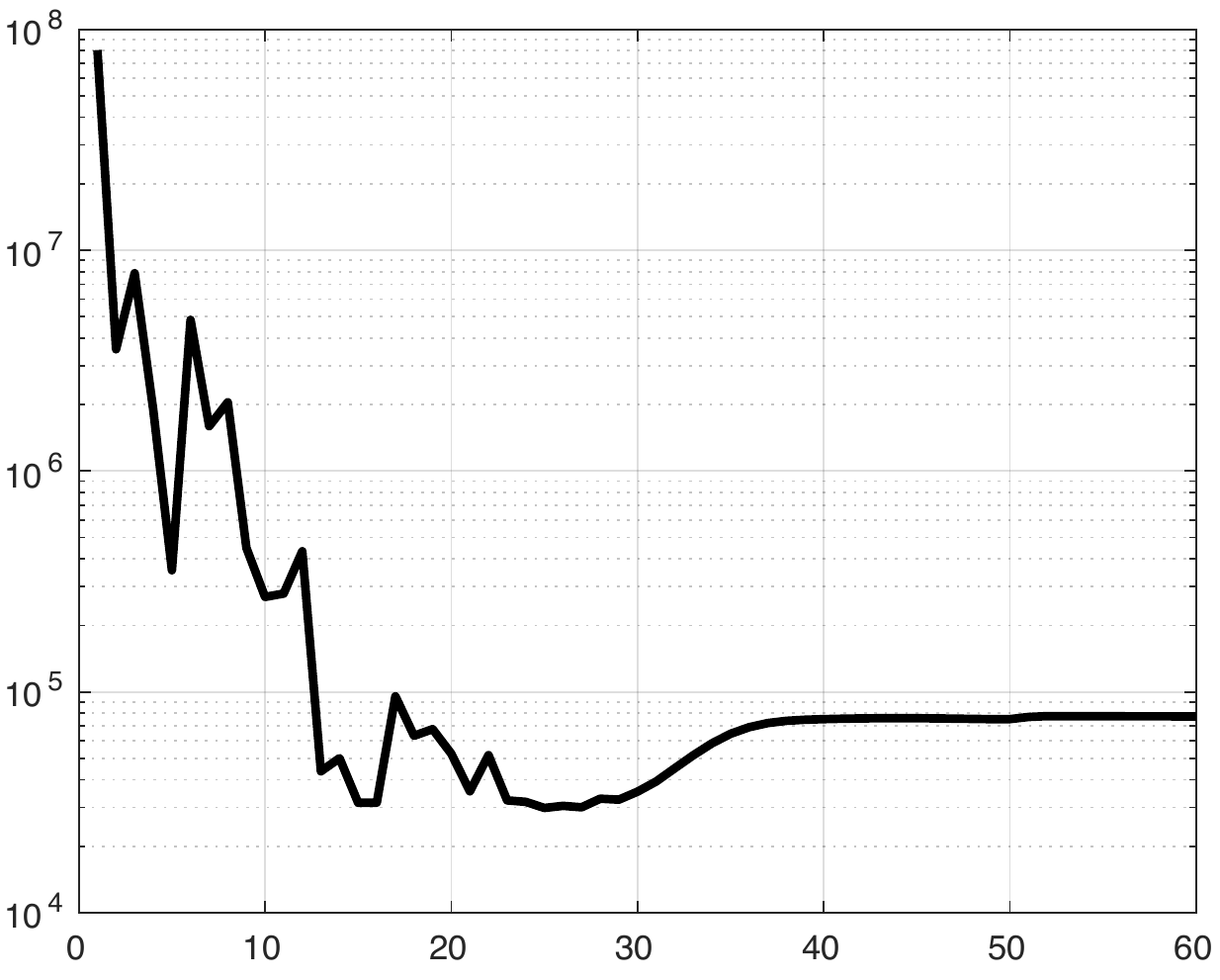}
		\caption{Objective with update.}
		\label{fig:obj}
	\end{minipage}
\end{figure*}
\subsection{Feature Descriptions}
Features adopted in this paper is shown as the following, each feature captures quite different information from images:
\begin{enumerate}
	\item CENTRIST\cite{wu2010centrist} stands for census transform histogram, is a holistic representation of images, which can be applied to capture the structural and textural properties from images. 
	\item 
	HOG\cite{dalal2005histograms} is based on oriented gradients, so it has great power to capture edge structures in images naturally. 
	\item 
	Color moment (CMT) \cite{yu2002color} represents the color distribution in images. The mathematical basis of this descriptor is that the color distribution can be represented efficiently by some low-order moments.
	\item 
	Local binary pattern captures the texture information from an image by computing the histogram of local binary patterns \cite{fei2004learning}. 
	\item 
	GIST \cite{potter1969recognition} is a global image feature. Gist features represent scene information from images well. It relates to the gradient information for different parts in an image, including scales and orientations. 
	\item Gabor, which is extracted by a Gabor filter and can do texture analysis and object detection in images. 
	\item Intensity (IT).
	Pixel intensity is the primary information stored within pixels, represents the densities of a certain pixel. 
\end{enumerate}

\subsection{Experiment Setup}
To evaluate the performance of our method, we compare our method with two subspace learning algorithms applied on single view: spectral clustering (SC)~\cite{ng2002spectral} and lower rank representation (LRR)~\cite{liu2012robust}, and five state-of-the-art multi-view methods including: pairwise co-regularized multi-view spectral clustering (P-CoReg)~\cite{kumar2011co}, centroid co-regularized multi-view spectral clustering (C-CoReg)~\cite{kumar2011co}, robust multi-view spectral clustering via low-rank and sparse decomposition (RMSC)~\cite{xia2014robust}, multi-view consensus graph clustering (MCGC)~\cite{zhan2018multiview}, and multi-view clustering via joint nonnegative matrix factorization (MNMF)~\cite{liu2013multi}.  

Detailed description about the methods mentioned above and the experiment process is as the following:
\begin{enumerate}
	\item Single view with SC.
	We run spectral clustering on each view-specific affinity matrix independently to get clustering results based on different features.
	\item Single view with LRR.
	We run LRR on each single feature set to get the low-rank subspace representation first, and then apply spectral clustering on each such representation we obtained. 
	\item P-CoReg, which makes the eigenvector matrix in standard spectral clustering method related to different views be close to each other, by employing pair-wise co-regularizers in the objective function. 
	\item C-CoReg, similar to P-CoReg, it regularizes the eigenvectors related to each specific view feature towards a consensus set. 
	\item RMSC, which is based on Markov chain method for clustering. 
	A shared low-rank transition probability matrix is used as a crucial input to the standard Markov chain method for clustering. 
	\item MCGC, where a consensus graph structure is learned by minimizing disagreement between diverse views and constraining the rank of the Laplacian matrix, it's able to obtain the cluster assignment directly from the consensus graph without any post-processing steps.
	\item MNMF, a joint nonnegative matrix factorization algorithm to regularize coefficient matrices learnt from different views towards a consensus, followed by $K$-means on the consensus matrix.
\end{enumerate}

\subsection{Experiment Results}
Fig.~\ref{fig:C} shows the subspace representation matrices $\mC$ obtained by different feature descriptors and the final consensus $\mC^*$ of the MSRC-v1 dataset. A good $\mC$ should have a clear block diagonal structure, since the data in the MSRC-v1 dataset is grouped by object classes. In Fig.~\ref{fig:C}, view-specific $\mC$ vary a lot from each other since they are capturing different characteristic from images. And for some of them, there is no obvious block structure, noise exists over the whole matrix. It is apparent that only relying on one single view-specific $\mC$ has a high chance to achieve poor result. But for the consensus $\mC^*$, the block diagonal structure is well-established and there is almost no noise off the diagonal blocks, which means each sample is well represented by the remaining data from the same object class, thus a great clustering result can be obtained based on it. We also utilize polynomial kernels to further improve the subspace learning performance. Fig.~\ref{fig:plot} shows the consensus $\mC^*$ obtained with linear kernel and polynomial kernel for the MSRC-v1 dataset, \textcolor{red}{\textbf{please zoom in to observe the details and differences}}. The block diagonal structure gets more recognizable with an appropriate polynomial kernel. Thus on complex datasets where nonlinear relationships exist, polynomial kernels can have superior performance compared to simple linear kernel. And from Fig.~\ref{fig:obj} we can see that the objective of the ADMM solver is converging as the iteration increases. 

Experiment results on six datasets are shown in Table~\ref{tab:1} and \ref{tab:2}, highest ACC and NMI for each dataset is highlighted. From the results it's not hard to conclude that certain view-specific $\mC$ cannot have a satisfying clustering performance, this may be caused by the fact that images from different clusters have great similarity in the characteristic captured by that view, for example, the color moment feature doesn't work well on the Handwritten dataset. But with multi-view clustering methods, independent feature sets are combined together to construct a view-consistent $\mC^*$, the clustering performance is greatly improved. What's more, in our proposed method, the consensus $\mC^*$ will not deviate from the optimal solution when a view-specific $\mC$ is not well learned, so our proposed method can achieve best clustering performance in most cases over the comparison methods. 

\begin{figure}\centering
	\includegraphics[width=\linewidth]{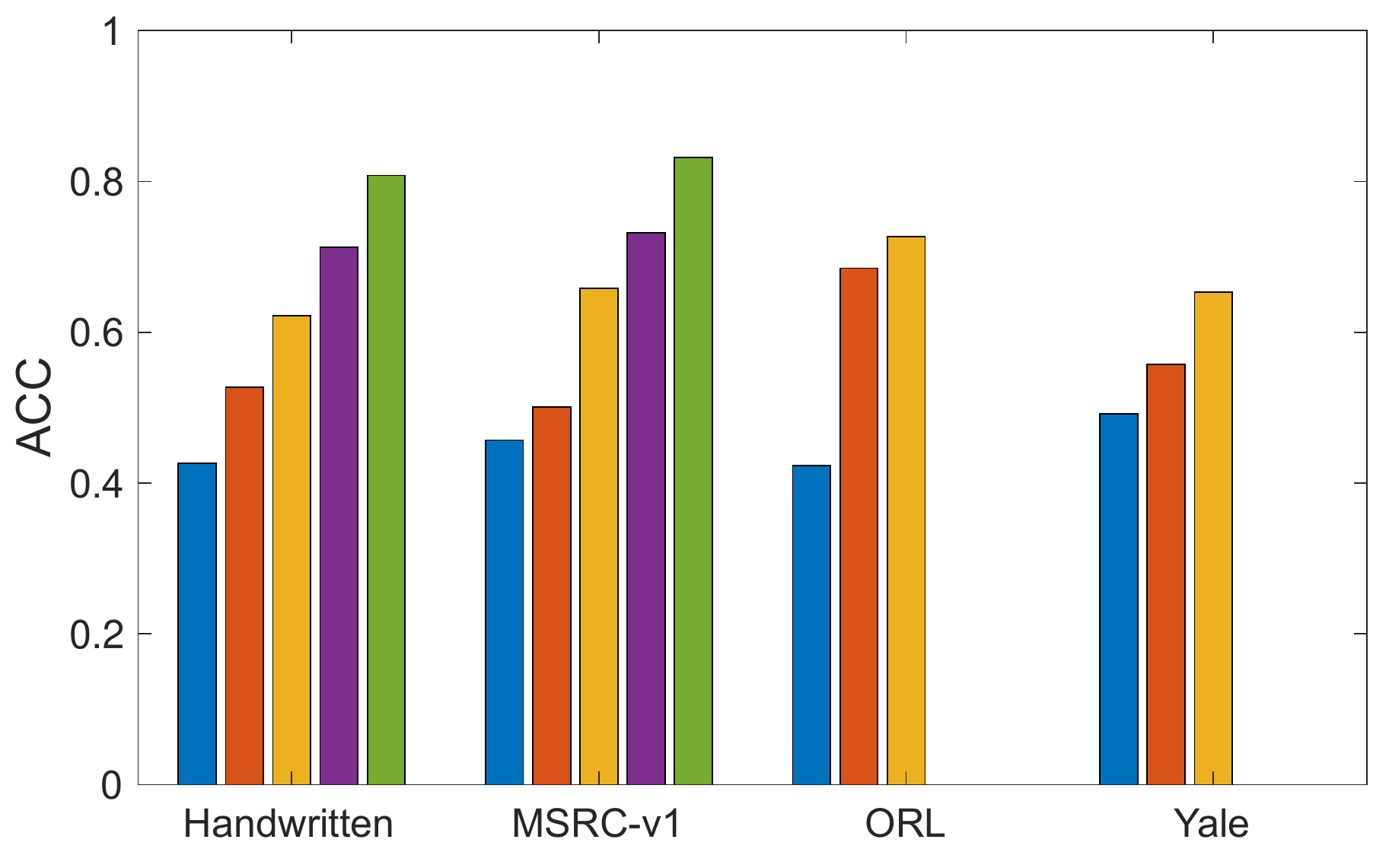}
	\caption{Accuracy comparison with increasing views, 1-5 views for Handwritten and MSRC-v1, 1-3 views for ORL and Yale.}
	\label{fig:acc_view}
\end{figure}

In addition, we investigate the performance of our method with varying parameter settings. There are three important parameters in our method: $\lambda$, $\gamma$, and $\theta$. We explore the effect of two parameters by fixing another. We present the results on MSRC-v1 dataset as Fig.~\ref{fig:cubic} demonstrates. From the figure we see that with the setting of $\lambda$ in the range of (0.001, 10), $\gamma$ in the range of (0.001, 10) and $\theta$ in the range of (0.001, 1), promising performance can be achieved. To show the advantage of combining multi-view feature sets in subspace clustering, we run our proposed method with increasing number of views on four datasets. The result is averaged on all the possible combinations of views, and for each combination we run the experiment 5 times. Comparison is shown in Fig.~\ref{fig:acc_view}. Apparently subspace clustering performance is improved significantly as number of views increases for all the datasets, since the data is described in a more comprehensive and extensive way. 


\section{Conclusion}
In this paper, we propose an \textit{Enriched Robust Multi-View Kernel Subspace Clustering}  model. Different from most existing multi-view clustering methods, our method obtains an enriched consensus affinity matrix from both the learning  and clustering stages. Besides, the proposed method extends linear space to kernel space to capture the nonlinear
structure hidden in the multi-view data. To optimize the objective with various constraints, we propose ADMM to obtain the optimal solution where in each step a closed solution is provided. 
Extensive experimental results on six benchmark datasets have
demonstrated the superiority of our method over several \textit{SOTA}
clustering methods.

{\small
\bibliographystyle{ieee_fullname}
\bibliography{references,egbib}
}

\section{Supplemental}
$\textbf{A}$. 
The detailed derivation of Eq.(\ref{eq:jsolution}) is as the following:
We denote the derivative of Eq.(\ref{eq:a}) regarding to $j$ as $H$, 

1) When $c^* > 0$:

a. If $j \geq c^*$: $H = \beta + \alpha + j - y = 0$, we can get $j = y - \alpha - \beta$, and with the assumption $j \geq c^*$, $y$ has to satisfy $y \geq \alpha + \beta + c^*$ to get this solution;

b. If $0 < j < c^*$: $H = \beta - \alpha + j - y = 0$, which leads to $j = y + \alpha - \beta$, and similarly due to the prerequisite $0 < j < c^*$, we should have $0 < y + \alpha - \beta < c^*$;

c. If $j \leq 0$: $H = -\beta - \alpha + j - y = 0$, we can get $j = y + \alpha + \beta$, so $y + \alpha + \beta \leq 0$ as well;

2) When $c^* < 0$:

a. If $j \geq 0$: $H = \beta + \alpha + j - y = 0$, we can get $j = y - \alpha - \beta$, and with the assumption $j \geq 0$, only when $y \geq \alpha + \beta$ we can get this solution;

b. If $c^* < j < 0$: $H = -\beta + \alpha + j - y = 0$, then we have $j = y - \alpha + \beta$, also $c^* < y - \alpha + \beta < 0$;

c. If $j \leq c^*$: $H = -\beta - \alpha + j - y = 0$, we can get $j = y + \alpha + \beta$, so when $y + \alpha + \beta \leq c^*$ we can get this solution.

$\textbf{B}$. 
The derivative details for Eq.(\ref{eq:csolution}) is as the following:

We suppose the optimal solution is $c^* = a_i$, and it's obvious that the subgradient of $|a_i - c^*|$ is any element in the interval of [-1, 1].

1) When $c^* > 0$:

To make the derivative of Eq.(\ref{c_start}) regarding to $c^*$ equal to 0, we should have
\begin{equation}
	\begin{aligned}
		-1 \leq 2\lambda(i-1) - 2\lambda(v-i) + \gamma q \leq 1,
	\end{aligned}
\end{equation}
which leads to 
\begin{equation}
	\begin{aligned}
		\frac{2v\lambda - \gamma q}{4\lambda} + \frac{1}{2} - \frac{1}{4\lambda} \leq i \leq \frac{2v\lambda - \gamma q}{4\lambda} + \frac{1}{2} + \frac{1}{4\lambda},
	\end{aligned}
\end{equation}
$i$ has to be an integer as an index, so $\left \lceil \frac{2v\lambda-\gamma q}{4\lambda} \right \rceil$ is an appropriate value for it. Also, an index $i$ has to be larger than 0, we should have $2v\lambda > \gamma q$, and since the assumption is $c^* > 0$, only when $a_{\left \lceil \frac{2v\lambda-\gamma q}{4\lambda} \right \rceil} > 0$ we can get this solution;

2) When $c^* > 0$:

To make the derivative of Eq.(\ref{c_start}) regarding to $c^*$ equal to 0, we should have
\begin{equation}
	\begin{aligned}
		-1 \leq 2\lambda(i-1) - 2\lambda(v-i) - \gamma q \leq 1,
	\end{aligned}
\end{equation}
which leads to 
\begin{equation}
	\begin{aligned}
		\frac{2v\lambda + \gamma q}{4\lambda} + \frac{1}{2} - \frac{1}{4\lambda} \leq i \leq \frac{2v\lambda + \gamma q}{4\lambda} + \frac{1}{2} + \frac{1}{4\lambda},
	\end{aligned}
\end{equation}
Again, $\left \lceil \frac{2v\lambda+\gamma q}{4\lambda} \right \rceil$ is an appropriate value for $i$, and the index should not exceed $v$, thus we have
\begin{equation}
	\begin{aligned}
		\frac{2v\lambda+\gamma q}{4\lambda} \leq v,
	\end{aligned}
\end{equation}
which leads to 
\begin{equation}
	\begin{aligned}
		2v\lambda \geq \gamma q,
	\end{aligned}
\end{equation}
similarly due to the assumption $c^* < 0$, only when $a_{\left \lceil \frac{2v\lambda+\gamma q}{4\lambda} \right \rceil} < 0$ we can get this solution.

When there is no such $a_{\left \lceil \frac{2v\lambda-\gamma q}{4\lambda} \right \rceil}$ or $a_{\left \lceil \frac{2v\lambda+\gamma q}{4\lambda} \right \rceil}$, it's easy to see $c^* = 0$ minimizes Eq.(\ref{c_start}). 

$\textbf{C}$.
Below are some figures that are not included in the main body of the paper due to space limitation:

The residual plot of the ADMM solver is presented in Fig.\ref{fig:res}, it's converging to 0 as the iteration increases:
\begin{figure}[h!]
	\centering
	\includegraphics[width=0.9\linewidth]{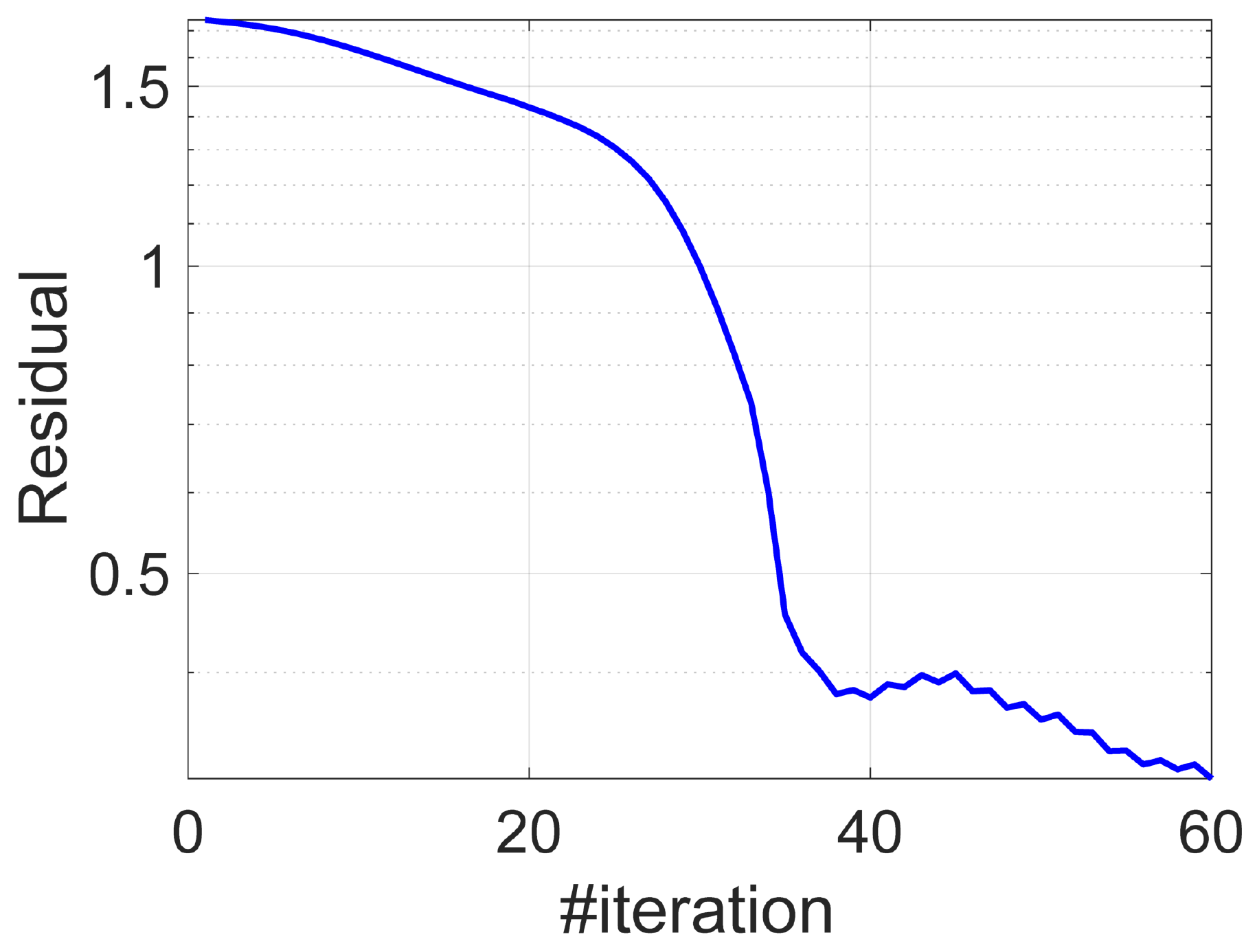}
	\caption{Residual with update.}
	\label{fig:res}
\end{figure}

To show the advantage of combining multi-view feature sets, in subspace clustering, we run our proposed method with increasing number of views on four datasets. Comparison of NMI is shown in Fig.\ref{fig:nmi_view}:
\begin{figure}[h!]
	\centering
	\includegraphics[width=0.9\linewidth]{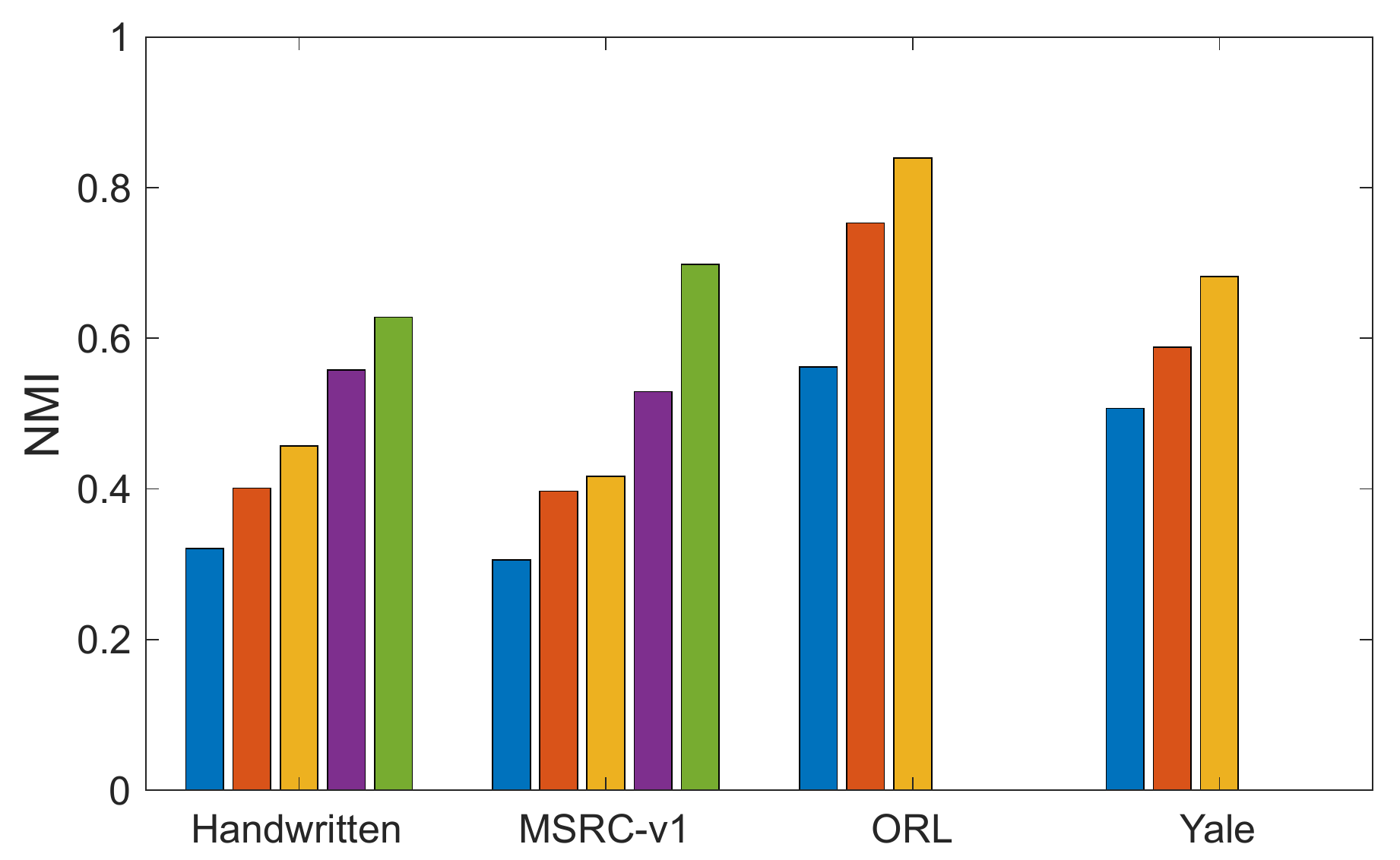}
	\caption{NMI comparison with increasing views, 1-5 views for Handwritten and MSRC-v1, 1-3 views for ORL and Yale.}
	\label{fig:nmi_view}
\end{figure}

\end{document}